# A Harmonic Potential Field Approach for Joint Planning & Control of a Rigid, Separable Nonholonomic, Mobile Robot

**Ahmad A. Masoud**

Electrical Engineering Department, KFUPM, P.O. Box 287, Dhaharan 31261, Saudi Arabia, masoud@kfupm.edu.sa

*Abstract* : The main objective of this paper is to provide a tool for performing path planning at the servo-level of a mobile robot. The ability to perform, in a provably-correct manner, such a complex task at the servo-level can lead to a large increase in the speed of operation, low energy consumption and high quality of response. Planning has been traditionally limited to the high level controller of a robot. The guidance velocity signal from this stage is usually converted to a control signal using what is known as an electronic speed controller (ESC). This paper demonstrates the ability of the harmonic potential field (HPF) approach to generate a provably-correct, constrained, well-behaved trajectory and control signal for a rigid, nonholonomic robot in a stationary, cluttered environment. It is shown that the HPF-based, servo-level planner can address a large number of challenges facing planning in a realistic situation. The suggested approach migrates the rich and provably-correct properties of the solution trajectories from an HPF planner to those of the robot. This is achieved using a synchronizing control signal whose aim is to align the velocity of the robot in its local coordinates, with that of the gradient of the HPF. The link between the two is made possible by representing the robot using what the paper terms "separable form." The context-sensitive and goal-oriented control signal used to steer the robot is demonstrated to be well-behaved and robust in the presence of actuator noise, saturation and uncertainty in the parameters. The approach is developed, proofs of correctness are provided and the capabilities of the scheme are demonstrated using simulation results.

Keywords: Harmonic potential, Navigation Control, Mobile robots, Motion planning,

**List of symbols**

| | |
|---|---|
| LBVP | Laplace Boundary Value Problem |
| FSR | Front Wheel Steered (car-like) Robot |
| DDR | Differential Drive Robot |
| UAV | Unmanned Aerial Vehicle |
| UGV | Unmanned Ground Vehicle |
| AUV | Autonomous Underwater Vehicle |
| AI | Artificial Intelligence |
| HLC | High Level Controller |
| LLC | Low Level Controller |
| HPF | Harmonic Potential Field |
| x | x coordinate of the robot in the global coordinates |
| y | y coordinate of the robot in the global coordinates |
| $\theta$ | orientation the robot in the global coordinates |
| $v$ | tangential speed of the robot in the local coordiantes |
| $\omega$ | angular speed of the robot in the local coordiantes |
| **P** | the global configuration vector ($\mathbf{P}=[x \; y \; \theta]^t$) |
| **X** | position vector of the robot in the global coordiantes ($\mathbf{X}=[x \; y]^t$) |
| $\boldsymbol{\lambda}$ | the local velocity vector ($\boldsymbol{\lambda}=[v \; \omega]^t$) |
| $\mathbf{X}_s$, $\mathbf{X}_T$ | start and target points respectively |
| **U** | the control vector |
| $\omega_R, \omega_L$ | right and left angular wheel velocity of the DDR robot |
| $\omega_h$ | angular speed of the rear wheels of the FSR robot |
| $\phi$ | steering angle of the front wheel of the FSR robot |
| **F** | local to global motion transformation stage, robot kinematic model |
| **Q** | control variable to local motion transformation, robot kinematic model |
| **F1**, **F2** | local to global motion transformation stage, robot dynamic model |
| **Q**$_D$ | control variable to local motion transformation, robot dynamic model |
| M | mass of the robot |
| I | inertia of the robot |
| W | wheel-to-wheel width of the DDR robot |
| L | steering wheel to driving wheel distance in the FSR robot |
| r | wheel radius |
| $T_L, T_R$ | left and right wheel torques, DDR robot |
| $\rho$ | spatial projection of the robot's trajectory |
| $\rho_n$ | spatial projection of the trajectory laid by the HPF planner |
| $\delta$ | deviation between $\rho$ and $\rho_n$ |
| $\delta_m$ | maximum deviation between $\rho$ and $\rho_n$ |
| $\dot{\mathbf{X}}_r(\mathbf{X})$ | reference velocity field |
| $\mathbf{S}$ | synchronizing control action |
| $\mathbf{S_D}$ | motion damping control action |
| $\Xi$ | Lyapunov function |
| $\mathbf{A}^+$ | pseudo inverse of $\mathbf{A}$ |
| K1, K2 | are positive constants. |
| KD1, KD2 | are positive constants. |
| $\eta_1, \eta_2$ | are non-negative functions. |
| $\Omega$ | the admissible space (workspace) of the robot |
| $\Gamma$ | boundary of $\Omega$, |
| $\mathbf{n}$ | unit vector normal to $\Gamma$, |
| O | set of forbidden regions |
| $\gamma(X)$ | cost of being at a location X. |
| $\mathbf{h}$ | unit vector in the target direction |
| V | scalar potential field |

## 1. Introduction

The integration of robotics in human activities has been an aggressively pursued goal for more than half a century. Robotics agents have demonstrated great success in tackling applications in structured and controlled environments. The momentum of this success is harvested through a push towards applications in unstructured, informationally impoverished environments [1]. This effort is scoring considerable success on almost all frontiers let that be unmanned aerial vehicles (UAVs), unmanned ground vehicles (UGVs), autonomous underwater vehicles (AUVs), etc. Although these robotics agents can be designed to assume any form, agents whose mechanical structures satisfy the rigidity constraints seem to be most prevalent and useful. Mobility is probably the most crucial issue in designing and operating a robot. Initially, mobility generation was tackled in one of two ways. Either the matter is considered as a control problem of a nonholonomic system or as a planning problem of a kinematic, single integrator system. When treated as a control problem [2-14] the issue of guidance is not considered. A reference trajectory is provided to the robot to convert into a control signal. For example, in [2] the motion of a unicycle is controlled using a constant tangential velocity and an angular velocity constructed from a sum of sinusoids. The control law guides the robot using only a scalar field whose unique maximum is situated on the target point. The same problem was tackled in [6] using a control law for the angular velocity constructed from the directional derivative of the scalar signal. In [9] a trajectory tracking nonlinear controller is suggested to adjust the angular velocity of a unicycle robot to make it coincide with that of a vector guidance field defined in free space. In [3] a survey of techniques for controlling nonholonmic systems that covers the state of art in the area until 1995 is provided. In [4] a neural net-based controller is suggested for enabling a dynamical mobile robot with uncertainty in the parameters to track a given trajectory. [5] investigated the tracking problem in nonholonomic robots and discussed issues relating to control design and system stabilization. A kinematic, car-like robot is used as a study case. [7] suggests a theoretical framework for the control of nonholonomic dynamical systems that is constructed from reduced order state equations. The framework is used to develop an approach based on geometric phases that is applied to the control of a knife edge moving on a plane or a surface In [8] a backstepping controller that avoids singularities and the need for repeated differentiations is suggested to enable a unicycle with uncertainty in the parameters to track a trajectory in free space. Modulated linear functions are used to curb the growth in the magnitude of the control signal. In [10] two approaches, the imbalanced Jacobian algorithm and the optimal control approach, are compared for controlling a unicycle robot when constrains on the control variables are present. In [11] the issue of singularity avoidance when synthesizing control for an over-actuated nonholonomic mobile robot is discussed. [12] proposed a Lyapunov-based method for controlling a unicycle robot in free space. The control signal is synthesized by multiplying the gradient of the lyapunov function by a tensor. In [13] the effect of uncertainty on the control of a kinematic, car-like robot is studied. A tutorial on the use of lie algebra in constructing controllers for nonholonomic systems is given in [14].

The planning approach taps mainly into artificial intelligence (AI) techniques. The aim is to convert a task, a description of the environment and a set of constraints, into a reference trajectory which the robot is required to follow [15-26]. In [15] the velocity space approach is suggested for planning a trajectory for a differential drive robot. The whole trajectory is assembled by joining local trajectory segments. Recursive application of lie bracket operation is used to guarantee that the whole trajectory is feasible for the robot to traverse. In [16] a method is suggested for laying a collision-free trajectory to a target point by a unicycle robot. Two behaviors are used in the trajectory construction: one to drive motion towards the target and the other to drive motion along the boundary of the detected obstacles. The idea of feasible velocity polygons is suggested to make sure that the generated path is feasible for traversal by the robot. In [17] the idea of parallel navigation is suggested to lay a feasible path for a unicycle to a moving target in a cluttered environment. In [18] the rapidly expanding random tree (RRT) approach is used to generate a feasible trajectory for a mobile robot that satisfies a given safety criterion. In [19] multi-criterial analysis is combined with RRT to enhance the quality of the selected trajectories. In [20] Dubins curves and sliding mode control are combined to lay a trajectory for a mobile robot using a scalar signal so that motion terminates at the signal's unique extremum. In [21] a generalized visibility graph approach that combines a group of Reed/Shepp paths is used to obtain a bounded curvature, minimum length trajectory which a nonholonomic mobile robot may use to reach a target in a cluttered environment. In [22] the concept of exponential navigation functions is utilized to generate a path from an initial configuration to a final one for a kinematic, car-like robot in free space. Some guidelines are given about how convex obstacles may be incorporated in the path generation process. In [23] an algorithm is suggested for constructing a vector field in a polygonal environment that is able to generate trajectories a unicycle can move along from anywhere in the work space to a target point. In [24] genetic algorithms combined with RRTs are used to generate a trajectory for a car-like, kinematic robot in a cluttered environment. The use of genetic algorithms aims at improving the quality of the solution obtained from the RRT algorithm. In [25] optimal control techniques are used to off-line generate feasible trajectories for nonholonomic robots in a cluttered environment. In [26] the random profile approach is used to generate, off-line, dynamically feasible trajectories for a differential drive robot working in a cluttered environment.

AI-based techniques rarely address the admissibility of the generated trajectory let alone the generation of the control signal that can actualize it. Successful mobility is a product of a context-sensitive, intelligent, goal-oriented, constrained process. This process operates as an interface between an operator and a robot. Its function is to interpret the commands and constraints on the process within the confines of the environment in which the robot is situated. The result is a sequence of control signal instructions that is able to actualize a task in a desired manner. This process is called a planner. Despite their diversity, planning methods [27-30] are divided into two classes. One class separates a planner into two modules (figure-1). One is called the high level controller (HLC) and the other is called the low level controller (LLC). The first is responsible for converting the command, constraints and environment feed into a desired behavior which the robot must find a way to actualize if the task is to be accomplished (a know-what-to-do guidance signal). The second module determines what actions the process actuators of motion should release in order to actualize the desired behavior (a know-how-to-do control signal).

Although, until this day, this division of role in building planners is widely accepted in the area of robotics, many researchers believe it is a source of several problems. It is well-known in practice that processes using the HLC-LLC paradigm are relatively slow and resource-hungry. Incompatibilities between the guidance and control signals could lead to unwanted artifacts in the behavior and undesirable control effort that consume too much energy or put too much strain on the actuators of the robot.

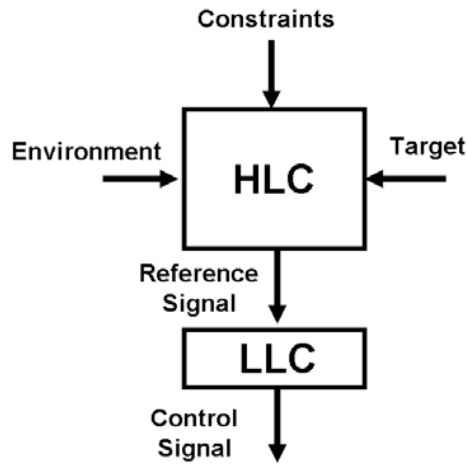

Figure-1: The two-tier, high-level low-level, planning paradigm

Jointly designing the guidance and control modules (figure-2) should yield a provably-correct and efficient planner compared to a design that treats the two modules separately. However, guaranteeing stability alone of a nonholonomic system [31-37] is an involved task. In [31] it was shown that a mobile robot has unstable internal dynamics when it moves backward. This instability does not get reflected in input-output presenations. In [32] it is shown that Lyapunov second method and the LaSalle invariance principle are sufficient for studying the stability of nonholonomic, mobile robots. In [33] it is proven that an attractive manifold does exist for guaranteeing stability of a nonholonomic system in a chained form. In [34] local and global uniform asymptotic stability of nonlinear, time varying switched systems with applications to mobile robots are studied. In [35] the existence of a smooth control law that can locally and(or) globally stabilize a nonholonomic system is proven provided certain geometric features are satisfied. A classical treatment of the stability issue in nonholonomic systems was provided by Brocket in [36]. In [37] it is argued that Nonholonomic cannot be asymptotically stabilized by a smooth pure state feedback. However smooth state feedback control laws can be designed to guarantee the global marginal stability of nonholonomic systems.

Simultaneous consideration of the guidance and control signals in the design of a planner, is a challenging and highly sought after task. An essential requirement to achieve this is the ability to enforce nonconvex constraints in the statespace of a nonlinear system. While limited success has been achieved in designing controllers that can incorporate simple avoidance regions with convex geometry in state-space [38,39], imposing nonconvex avoidance regions in the state-space of a dynamical system [40,41] is difficult. The task is further complicated when state-space constraints have to be implemented along with constraints in the control space as is the case with dynamical, nonholonomic systems.

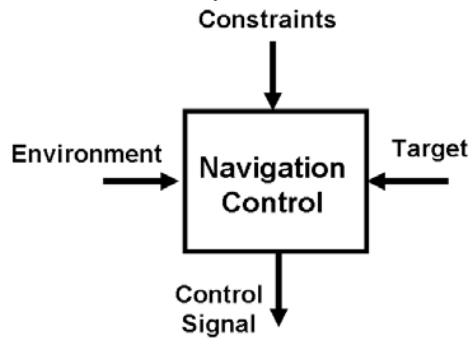

Figure-2: The joint guidance and control planning paradigm

A robust and versatile planner that enables a robot to operate in a realistic environment is expected to satisfy a long list of requirements. For example, it should be able to impose a variety of constraints on motion as well as accept different formats of the data describing the environment. Simplicity, robustness, ease of implementation and tuning are all highly desirable. Added to the above is the planner's ability to generate a rich variety of behaviors which the agent may need to execute during deployment. Co-existing with other robots in an environment is also important. The requirements in this long list are highly unlikely to be met if the planning and control signals are to be simultaneously designed.

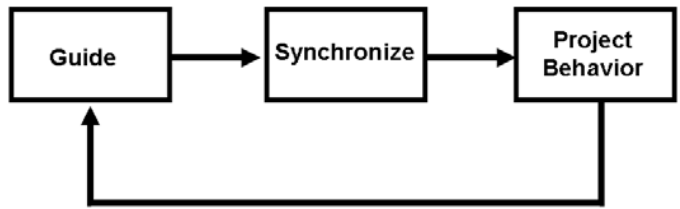

Figure-3: The suggested velocity synchronizing paradigm for planning

Instead of using the relatively simple two-tier approach to planner design or the excessively difficult joint state-space control-space approach (both approaches use an open loop design strategy), this paper adopts a middle ground that relies on a close loop design strategy (figure-3). The suggested approach starts with a well-behaved planning guidance signal that densely covers the workspace of the robot with solution trajectories (a goal-seeking planning action). The planning signal generates only kinematic trajectories that satisfy the desired constraints on behavior. Although, at first glance, the planning signal may appear to totally disregard the dynamics of the robot. The fact is that dynamics are being implicitly considered by generating well-behaved kinematic solution trajectories. The trajectories are well-behaved in the sense that a nonlinear inverse [42,43] exists that can map the planning signal into a control signal. The planning signal assumes the form of a desired, differential, vector increment along which the dynamic trajectory is expected to move. This velocity reference direction is followed by a synchronizing stage. The synchronizing stage generates the control signal that progressively aligns the differential motion of the state with the reference provided by the planner. Regardless of the current location of the state, synchronization is continuously attempted until the actual velocity converges to the reference velocity. Once the two differential velocity elements coincide with each other, the kinematic and dynamic trajectories become identical. This in effect causes all the properties encoded in the guidance trajectory to migrate to the robot's dynamical trajectory.

Modeling of mobile robots is an extensively investigated subject [44,45]. However, in this paper, the above strategy utilizes what the author terms "separable models" for rigid unmanned ground vehicles (UGVs). A UGV is called separable if its model can be divided into two independent stages. The first stage ties the control variables to the motion variables in the local coordinates of the robot. The second stage transforms the local motion variables into global motion variables. It ought to be mentioned that the above class admits a wide variety of UGVs, including the two most popular forms: differential drive and car-like UGVs.

The potentnial field approach to motion planning demonstrated a promising potential for use in navigating nonholonomic mobile robots [46-52]. In [44] potential fields are combined with the modified Newton optimization method to generate a trajectory for a kinematic, unicycle robot in a cluttered environment. In [47] the idea of ring sphere potential is suggested for navigating a kinematic unicycle robot in a cluttered environment. The idea was further developed by the authors in [48]. In [49,50] potential fields are coupled with sliding mode control to navigate a dynamical robot in an environment with obstacles. In [51] a procedure is suggested for combining potential fields and sensor data to safely move a wheel chair in an environment with obstacles. A potential field-based method that can tolerate uncertainty in the robots parameters is suggested for controlling the motion of a kinematic, nonholonomic robot in free space.

In particular, the harmonic potential field (HPF) planning approach [53-58] seems to be most suited for generating the well-behaved planning signal needed for realizing the suggested planning paradigm. The HPF guidance field satisfies most, if not all, of the properties desired in a planner that fits the above framework. A synchronizing controller that capitulate on the properties of the two-stage model and the HPF method is suggested for aligning the velocity of the UGV with the reference velocity of the planner. The approach is developed for both the kinematic and kinodynamic cases. Basic proofs of correctness are provided along with extensive simulation results to demonstrate the ability of the suggested scheme in dealing with realistic situations.

## 2. HPF Planners: A Background

This section provides a brief background of the HPF approach. It first presents the different variants of the HPF approach to planning and the situations in which each is used. The means available for computing the HPF filed are then briefly discussed.

## 2.1 HPF-based Planning Techniques: An Overview

The HPF approach is an excellent goal-seeking planner. It works by inducing a dense collective of guidance vectors (figure-4) on a $C^\infty$ surface (the harmonic potential) covering the admissible space of the robot ($\Omega$). A group structure is then evolved on this collective to generate a macro-template encoding the guidance information the process needs to execute. The action selection mechanism the approach utilizes for generating the field structure is in conformity with the artificial life (AL) method [59]. The HPF approach offers a solution to the local minima problem faced by the potential field approach Khatib suggested in [60]. It was simultaneously and independently proposed by several researchers [53-58] of whom the work of Sato in 1987 may be regarded as the first on the subject [57]. An HPF is generated using a Laplace boundary value problem (BVP) configured using a properly chosen set of boundary conditions (1). There are several settings one may use for a Laplce BVP (LBVP) in order to generate a navigation potential [53-58]. Each one of these settings possesses its own, distinct, topological properties [56]. An example is shown below (1) of an LBVP that is configured using the homogeneous Neumann boundary conditions. The setting encodes region avoidance constraints and target location

solve $\qquad \nabla^2 V(\mathbf{X}) \equiv 0 \qquad \mathbf{X} \in \Omega$ (1)

subject to $V(\mathbf{X}_S) = 1$, $V(\mathbf{X}_T) = 0$, and $\dfrac{\partial V}{\partial \mathbf{n}} = 0$ at $\mathbf{X} = \Gamma$,

where $\Omega$ is the workspace, $\Gamma$ is its boundary, $\mathbf{n}$ is a unit vector normal to $\Gamma$, $X_s$ is the start point, and $X_T$ is the target point. The trajectory ($\mathbf{X}(t)$) is generated using the dynamical system

$$\dot{\mathbf{X}} = -\nabla V(\mathbf{X}) \qquad \mathbf{X}(0) = \mathbf{X}_0 \in \Omega \qquad (2)$$

The trajectory is guaranteed to

1- $\lim_{t \to \infty} \mathbf{X}(t) \to \mathbf{X}_T$

2- $\mathbf{X}(t) \in \Omega \qquad \forall \mathbf{t}$

Equation (3) provides a LBVP similar to (1) that adds target orientation to the set of encoded features. The LBVP is:
Solve

$$\nabla^2 V(\mathbf{X}) \equiv 0 \ \mathbf{X} \in \Omega \qquad (3)$$

subject to $\qquad V(\mathbf{X}_S) = 1, V(\mathbf{X}_T) = 0, V(\mathbf{X}_T + \epsilon \cdot \mathbf{h}) = 1$, $\dfrac{\partial V}{\partial \mathbf{n}} = 0$ at $\mathbf{X} = \Gamma$,

where $1 \gg \epsilon > 0$ and $\mathbf{h}$ is a unit vector in the target direction.

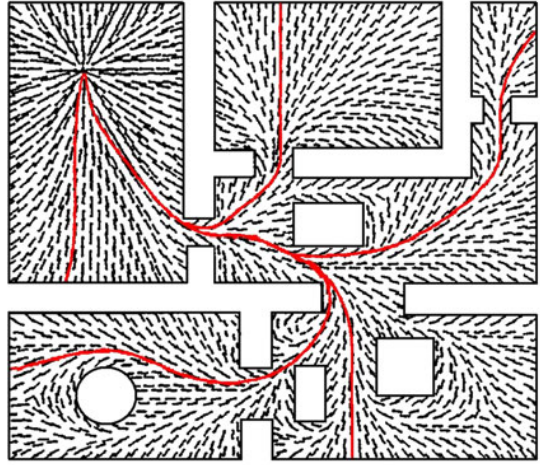

Figure-4: The reference velocity field from an HPF.

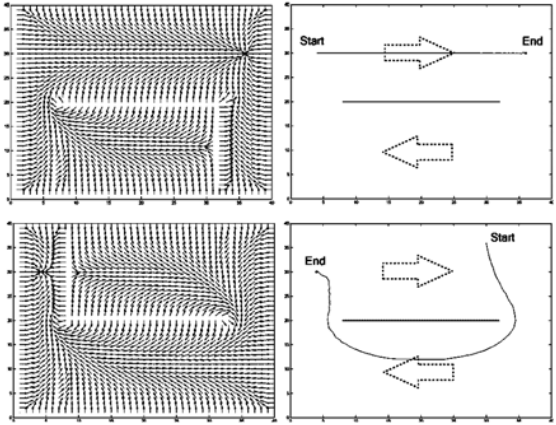

Figure-5: Directional & regional avoidance constraints

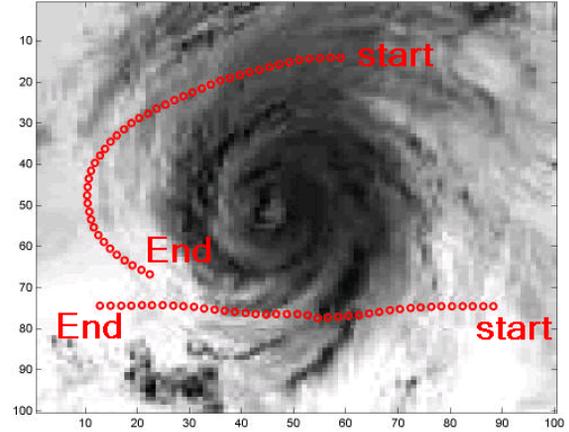

Figure-6: Planning in non divisible environments

Harmonic functions have many useful properties [62] for motion planning. Most notably, a harmonic potential is also a Morse function [63] and a general form of the navigation function suggested in [64]. The HPF approach may be configured to operate in a model-based and/or sensor-based mode. It can also be made to accommodate a variety of constraints [61]. It ought to be mentioned that the HPF approach is only a special case of a much larger class of planners called: evolutionary, pde-ode motion planners [75]. Many variants of the settings in (1) and (3) were later proposed to extend the capabilities of the HPF approach. For example, it is demonstrated that the approach can be used for planning in complex unknown environments [75] relying on local sensing only. The HPF approach can incorporate directional constraints along with regional avoidance constraints [61] in a provably-correct manner to plan a path to a target point. In figure-5 a rectangular environment is divided by a line obstacle into an upper and a lower region. Opposite directional constraints are to be enforced by the planner in each region along with the obstacle avoidance constraints. In the first case the HPF planner selected the straightforward, constraint-satisfying solution of driving motion along a straight line to the target. When the start and target were interchanged, the planner abandoned the simple solution and generated a more involved trajectory in order to jointly satisfy the directional and obstacle avoidance constraints. The navigation potential may be generated using the BVP in (4)

$$\text{jointly solve} \quad \nabla^2 V(\mathbf{X}) \equiv 0 \quad \mathbf{X} \in \Omega - \Omega' \quad (4)$$
$$\text{and} \quad \nabla \cdot [\Sigma(\mathbf{X}) V(\mathbf{X})] = 0 \quad \mathbf{X} \in \Omega'$$
$$\text{subject to} \quad V(\mathbf{X}) = 1 \text{ at } \mathbf{X} = \Gamma \text{ and } V(\mathbf{X}_T) = 0$$

$$\text{where} \quad \Sigma(\mathbf{X}) = \begin{bmatrix} \sigma(\mathbf{X}) & 0 & \dots & 0 \\ 0 & \sigma(\mathbf{X}) & \dots & 0 \\ & & \ddots & \\ 0 & 0 & \dots & \sigma(\mathbf{X}) \end{bmatrix}, \quad \sigma(\mathbf{X}) = \begin{bmatrix} \sigma_f & -\nabla V(\mathbf{X})^t \Lambda(\mathbf{X}) > 0 \\ \sigma_b & -\nabla V(\mathbf{X})^t \Lambda(\mathbf{X}) \leq 0 \end{bmatrix} \quad (5)$$

where $\Omega'$ is a subset of $\Omega$ on which the directional constraints are defined, $\nabla V(X)$ is the gradient field of the HPF, $\Lambda(X)$ is a vector field whose domain is $\Omega'$. It defines the desired directional constraints. A provably correct trajectory to the target that enforces both regional avoidance and directional constraints may be simply obtained using the gradient dynamical system in (2). The HPF approach may be modified to take into consideration inherent ambiguity [65,66] that prevents the partitioning of an environment into admissible and forbidden regions. In figure-6 the HPF-based planner is required to generate a safe path from a start point to a target point in a turbulent environment. The environment is described using the magnitude of turbulence (dark regions indicate high turbulence, light regions indicate low level of turbulence). As can be seen, the HPF-based planner managed to lay a smooth trajectory of reasonable length from the start to the target points that passes mostly through low turbulence regions. The approach is called the gamma-harmonic potential field (GHPF). The BVP that generates the navigation potential for this case (6) is solve $\quad\quad\quad\quad\quad\quad\quad\quad \nabla \cdot (\gamma(\mathbf{X}) \nabla V(\mathbf{X})) \equiv 0 \quad\quad\quad\quad \mathbf{X} \in \Omega$ (6)

subject to $\quad\quad\quad\quad\quad\quad\quad V(\mathbf{X}_S) = 1, \ V(\mathbf{X}_T) = 0$

where $\gamma(X)$ is the cost of being at a location X. A provably correct path that avoids definite threat regions ($\gamma(\mathbf{X})=0$) and converge to the target may be generated using the gradient dynamical system in (2). The HPF-based approach may be easily modified to take advantage of a drift field that maybe present in an environment [67] in a manner that suits planning for energy exhaustive missions. It was also demonstrated in [68] that the HPF approach can work with integrated navigation systems that can efficiently function in a real-life situation. An HPF-based, decentralized, multi-agent approach was suggested in [69].

## 2.2 Means for computing an HPF

There is a wide spectrum of techniques that may be used to compute a harmonic potential. These techniques posses high diversity that spans classical, modern, hardware and software methods. For example, there is a large family of numerical methods that may be used to generate an HPF. The theory behind these techniques is well studied and efficient implementations are available via open source and off-the-shelf numerical packages for 2D and 3D environments. Some of the major approaches in this area are: the finite difference methods (FDM) [76] which is a basic and robust numerical method that may be used to solve all the variants of the HPF approach in a reasonably short time (less than a second depending how fast the computational platform is). The finite element method (FEM) provides a faster solution compared to that generated by an FDM [77]. However, the boundary element method (BEM) is the fastest and can generate a continuous, 2D harmonic potential in realtime (few milliseconds).

The hardware techniques that may be used to compute an HPF are as diverse as the software ones. For example, HPF-based neural nets demonstrated their ability to navigate a mobile robot in realtime [79,80]. FPGAs were used in [81] to generate an HPF on a 50x50 grid in less than 100 micro-seconds. Analog VLSI's manufactured in the mid nineties [82,83] were able to compute an HPF in few milliseconds. With the advances in analog VLSI technology, a chip developed for such a purpose nowadays does have the ability to generate an HPF in few micro-seconds.

Another source of the computational efficiency of the HPF approach is a result of its nature. Whole domain computation is not needed when a new component of the environment is introduced to an existing navigation HPF. The field has to only be recomputed in the local neighborhood of the newly introduced component. This property was clearly demonstrated in [75]. Not only it made it possible to use the HPF in sensor-based navigation, it also made the computation very efficient. Using this feature, Sato demonstrated in [84] the ability to efficiently compute an HPF in 4D space using only a basic numerical technique such as FDM.

## 3. The Two-stage Model

Models that fit most wheeled mobile robots were discussed in [45]. Here a model for what this paper terms "separable mobile robot" is considered. A separable robot is a robot whose model can be expressed as a cascade of two stages (figure-7). The first stage captures the actuation of motion in the local coordinates of the robot (7). It ties the control vector $\mathbf{U}$ to the local velocity vector $\boldsymbol{\lambda}$. The second stage transforms the local velocity vector $\boldsymbol{\lambda}$ into the global velocity vector $\dot{\mathbf{P}}$. This setting is in accordance with the work in [45]. It lends itself to the natural causality of action generation. Moreover, it is of key importance to constructing the planner suggested in this paper.

The kinematic equations of a separable mobile robot are represented as

$$\begin{aligned} \dot{\mathbf{P}} &= \mathbf{F}(\mathbf{P})\lambda \\ \lambda &= \mathbf{Q}(\mathbf{U}) \end{aligned}. \tag{7}$$

For a differential drive robot (DDR, figure-8A), the kinematic equations of motion (8) are

$$\begin{bmatrix} \dot{x} \\ \dot{y} \\ \dot{\theta} \end{bmatrix} = \begin{bmatrix} \cos(\theta) & 0 \\ \sin(\theta) & 0 \\ 0 & 1 \end{bmatrix} \begin{bmatrix} v \\ \omega \end{bmatrix}, \quad \begin{bmatrix} v \\ \omega \end{bmatrix} = \begin{bmatrix} \dfrac{r}{2} & \dfrac{r}{2} \\ \dfrac{r}{W} & \dfrac{-r}{W} \end{bmatrix} \begin{bmatrix} \omega_R \\ \omega_L \end{bmatrix}, \quad \begin{bmatrix} \omega_R \\ \omega_L \end{bmatrix} = \begin{bmatrix} \dfrac{1}{r} & \dfrac{W}{2r} \\ \dfrac{1}{r} & \dfrac{-W}{2r} \end{bmatrix} \begin{bmatrix} v \\ \omega \end{bmatrix} \tag{8}$$

where $\mathbf{P}=[x\ y\ \theta]^t$, $\boldsymbol{\lambda}=[v\ \omega]^t$, $\mathbf{U}=[\omega_R\ \omega_L]^t$, r is the radius of the robot's wheels, W is the width of the robot, $\omega_R$ and $\omega_L$ are the angular speeds of the right and left wheels of the robot respectively, $v$ is the tangential velocity of the robot and $\omega$ is its angular speed.

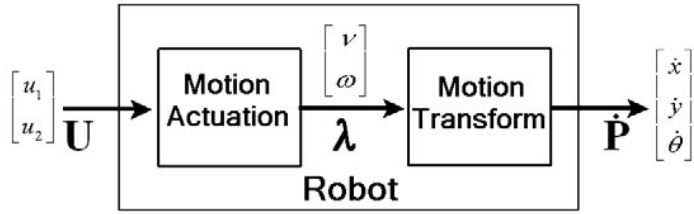

Figure-7: Two-stage model of a rigid, nonholonomic robot.

For the car-like, front wheel steered robot (FSR, figure-8B), the equations describing the kinematics of motion (9) are

$$\begin{bmatrix} \dot{x} \\ \dot{y} \\ \dot{\theta} \end{bmatrix} = \begin{bmatrix} \cos(\theta) & 0 \\ \sin(\theta) & 0 \\ 0 & 1 \end{bmatrix} \begin{bmatrix} v \\ \omega \end{bmatrix}, \quad \begin{bmatrix} v \\ \omega \end{bmatrix} = \begin{bmatrix} r \cdot \omega_h \\ r \cdot \omega_h \cdot \dfrac{\tan(\phi)}{L} \end{bmatrix}, \quad \begin{bmatrix} \omega_h \\ \phi \end{bmatrix} = \begin{bmatrix} v/r \\ \tan^{-1}((\omega \cdot L)/v) \end{bmatrix} \qquad (9)$$

where L is the normal distance between the center of the front wheel and the line connecting the rear wheels, $\omega_h$ is the angular speed of the rear wheels, and $\phi$ is the steering angle of the front wheel ($\pi/2 > \phi > -\pi/2$).

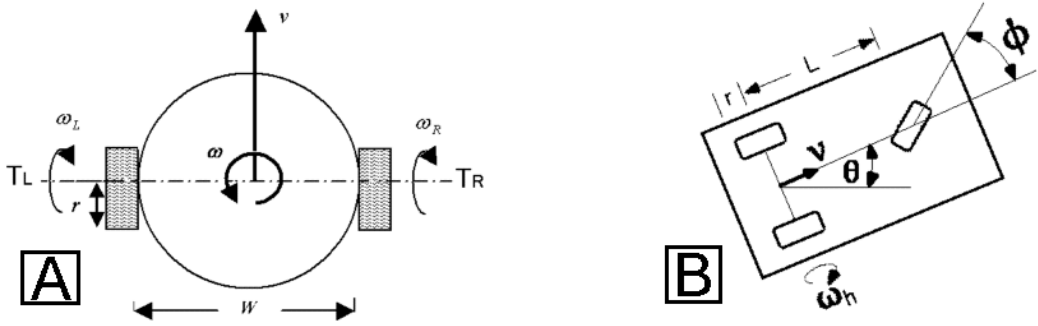

Figure-8: A-DDR mobile robot, B-FSR mobile robot.

The dynamics of a separable robot may be placed in the form (10)

$$\begin{aligned} \ddot{\mathbf{P}} &= \mathbf{F1}(\mathbf{P},\dot{\mathbf{P}})\lambda + \mathbf{F2}(\mathbf{P},\dot{\mathbf{P}})\dot{\lambda} \\ \dot{\lambda} &= \mathbf{Q_D}(\lambda,\mathbf{U}) \end{aligned} \qquad (10)$$

Under the assumption of dynamic equilibrium, the equation of motion for the first stage (11) of the DDR (local to global motion transformation) may be obtained by differentiating the velocity equation in (9)

$$\begin{bmatrix} \ddot{x} \\ \ddot{y} \\ \ddot{\theta} \end{bmatrix} = \begin{bmatrix} \cos(\theta) & 0 \\ \sin(\theta) & 0 \\ 0 & 1 \end{bmatrix} \begin{bmatrix} \dot{v} \\ \dot{\omega} \end{bmatrix} + \begin{bmatrix} -\sin(\theta)\dot{\theta} & 0 \\ \cos(\theta)\dot{\theta} & 0 \\ 0 & 0 \end{bmatrix} \begin{bmatrix} v \\ \omega \end{bmatrix}. \qquad (11)$$

The second actuation stage (12) is derived using Lagrange dynamics in the natural coordinates of the robot,

$$\begin{bmatrix} \dot{v} \\ \dot{\omega} \end{bmatrix} = \dfrac{1}{M} \begin{bmatrix} \dfrac{1}{r} & \dfrac{1}{r} \\ \dfrac{-4 \cdot r}{W^3} & \dfrac{4 \cdot r}{W^3} \end{bmatrix} \begin{bmatrix} T_R \\ T_L \end{bmatrix} \qquad (12)$$

Where M is the mass of the robot, $T_R$ and $T_L$ are the right and left torques applied to the wheels.

## 4. Problem Statement

The suggested approach operates by migrating the provably-correct properties of a kinematic, HPF planner to the control signal. Proving the utility of the suggested planner requires showing that:
1- Stability is guaranteed.
2- Motion of the robot will align (become in-phase) with the guidance signal from the planner. This will result in migrating the provably-correct navigation properties from the guidance field to the control field.
3- The deviation of the dynamic path from the kinematic path set by the harmonic planner is finite and controllable.

With no loss of generality, consider a harmonic potential planning field with the basic capabilities of convergence and obstacle avoidance. The gradient of the potential is used to construct the gradient dynamical system in (2) to generate the obstacle-free, convergent, kinematic solution trajectory

such that
$$\dot{\mathbf{X}} = -\nabla V(\mathbf{X}) \qquad \mathbf{X}(0) = \mathbf{X}_0$$
$$\lim_{t\to\infty} \begin{bmatrix} x(t) \\ y(t) \\ \theta(t) \end{bmatrix} \to \begin{bmatrix} x_T \\ y_T \\ \theta_T \end{bmatrix} \qquad (13)$$
and
$$\mathbf{X}(t) \cap O \equiv \varnothing \qquad \forall t$$

where, $\mathbf{X}=[x\ y]^t$ and O is the set of forbidden regions.

Let a reference velocity function be defined as $\dot{\mathbf{X}}_r(\mathbf{X}) = -\nabla V(\mathbf{X})$.

For both the kinematic and the dynamic robot models in (7,10), it is required to find a control signal ($\mathbf{U}(t)$) such that when the guidance system in (13) is used, the robot will satisfy (14)

$$\lim_{t\to\infty} \mathbf{P}(t) \to \mathbf{P}_T$$
$$\left|\dot{\mathbf{X}}(t) - \dot{\mathbf{X}}_r(t)\right| < \varepsilon, \qquad \forall\, t > T_\varepsilon, \qquad (14)$$
$$\delta(t) < \delta_m,$$

where $\delta_m$ is the controllable maximum deviation between the path laid by the HPF planner and the actual path traversed by the robot and $T_\varepsilon$ is a finite and positive number.

## 5. The Suggested Controller

There are more than one reason why the HPF approach can be adapted for generating efficient navigation control signals for rigid robots. The most important one has to do with the smoothness of the solution trajectories the HPF approach generates. These trajectories are $C^\infty$. According to [45], a trajectory being at least $C^2$ guarantees a realizability for most mobile robots. Therefore, analytic trajectories will suit most, if not all, mobile robots. The second reason has to do with the manner in which the approach functions. The HPF approach breaks down the global task into a dense collective of local reference signals that are spread all over the workspace of the robot ($\Omega$). A reference signal is merely an N-D vector with a base that is immobilized to a certain point $\mathbf{X}_i$ in the work space(figure-9). In an HPF approach this vector is usually selected as the negative gradient of a properly conditioned harmonic potential field. These vectors are used as a reference velocity field. If at a certain point in space the velocity of the state of the robot is made to coincide with the reference velocity, its behavior will start mimicking the trajectory from the HPF planner (figure-10).

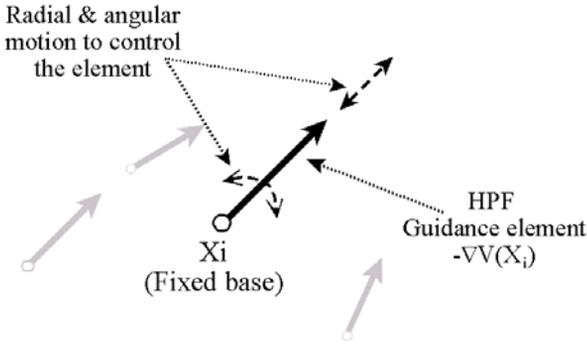
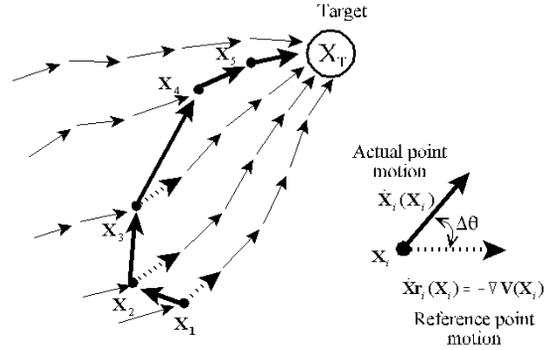

Figure-9: differential guidance element of an HPF planner     Figure-10: migration of the HPF guidance properties to a robot

The above may be implemented as follows. The current reference velocity signal is compared to the actual velocity of the robot and a synchronizing control action whose aim is to align the two is generated (figure-11). The robot will only use the first control instruction in the sequence associated with a solution trajectory and discard the remaining ones. The process is continuously repeated until the reference speed and the actual one coincide with each other. This paper demonstrates that the above paradigm does provide a good basis for building a provably-correct nonholonomic motion planner for rigid robots. The nature of the paradigm does not limit the construction of planners for planar robots. It also makes it possible to deal with three dimensional even N-D spaces [70,71].

For the kinematic case, the control signal (figure-12) is generated as follows. First, the synchronizing signal $\mathbf{S}=[s_1\ s_2]^t$ (15) is generated

$$\mathbf{S} = \begin{bmatrix} s_1 \\ s_2 \end{bmatrix} = \begin{bmatrix} K1 \cdot \left|-\nabla V\right| \cdot \cos(\arg(-\nabla V) - \theta) \\ K2 \cdot (\arg(-\nabla V) - \theta) \end{bmatrix} \qquad (15)$$

where K1 and K2 are positive constants. The inverse of the actuation stage ($\mathbf{Q}^{-1}$) is then applied to generate the control signal $\mathbf{U}$
$$\mathbf{U} = \mathbf{Q}^{-1}(\mathbf{S}). \qquad (16)$$

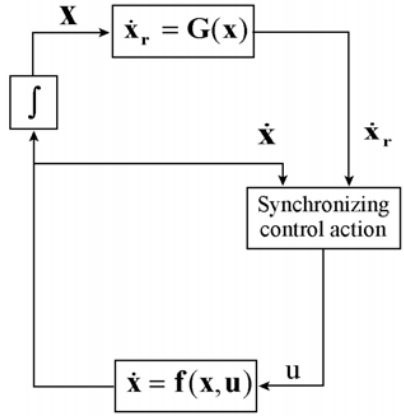

Figure-11: making a nonlinear system emulate the dynamics of an HPF-based gradient system

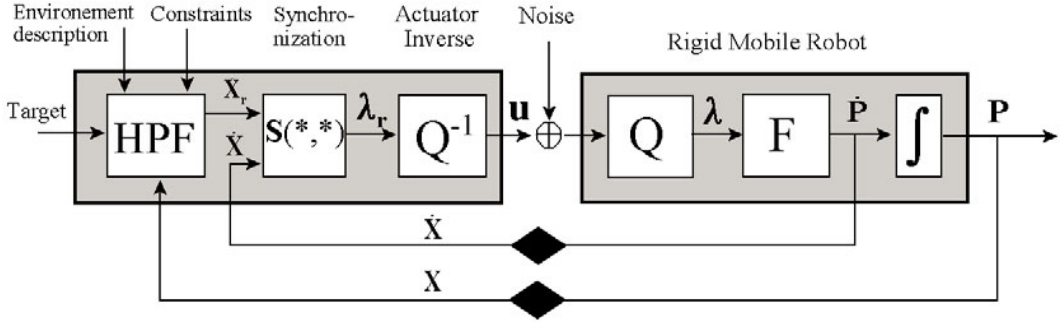

Figure-12: suggested planner for kinematic, separable mobile robots

For the DDR robot, the control signal is (17)
$$\mathbf{U} = \begin{bmatrix} \omega_R \\ \omega_L \end{bmatrix} = \begin{bmatrix} \frac{1}{r} & \frac{W}{2r} \\ \frac{1}{r} & \frac{-W}{2r} \end{bmatrix} \begin{bmatrix} s_1 \\ s_2 \end{bmatrix} \tag{17}$$

For the FSR robot, the control signal is (18)
$$\mathbf{U} = \begin{bmatrix} \omega_h \\ \phi \end{bmatrix} = \begin{bmatrix} s_1 \\ \tan^{-1}(s_2 / (L \cdot s_1)) \end{bmatrix} \tag{18}$$

For the dynamic case the following synchronizing, generalized, force signal (19) is used
$$\mathbf{S} = \begin{bmatrix} s_1 \\ s_2 \end{bmatrix} = \begin{bmatrix} K1 \cdot (|-\nabla V| - \nu) \cdot \cos(\arg(-\nabla V) - \theta) \\ K2 \cdot (\arg(-\nabla V) - \theta) \end{bmatrix} \tag{19}$$

where $\nu$ is the tangential speed of the robot, K1 and K2 are positive constants. A damping force $\mathbf{S_D}$ (20) is added to the synchronizing force to ensure stability in the local coordinates of the robot

$$\mathbf{S_D} = \begin{bmatrix} KD1 \cdot \eta_1(-\nabla V, \dot{\mathbf{X}}) & 0 \\ 0 & KD2 \cdot \eta_2(-\nabla V, \dot{\mathbf{X}}) \end{bmatrix} \begin{bmatrix} \nu \\ \omega \end{bmatrix} \tag{20}$$

where $\eta_1$ and $\eta_2$ are positive functions. For the DDR robot, $\eta_1$ and $\eta_2$ may be selected in two ways (21)
$$\eta_1 = 1, \quad \eta_2 = 1,$$
$$\text{or} \quad \eta_1 = 1 - \frac{\dot{\mathbf{X}}^t(-\nabla V)}{\|\dot{\mathbf{X}}\| |-\nabla V|} = 1 - \cos(\arg(-\nabla \mathbf{V}) - \theta), \quad \eta_2 = 1. \tag{21}$$

where X is the position of the robot in the x-y plane (X=[x y]$^t$) The second choice of $\eta_1$ is expected to produce a more agile response than the first choice. This is because motion damping will be zero when the velocity of the robot is in aim with the reference velocity of the gradient field. KD1 and KD2 are positive constants ($K_\lambda$=[KD1 KD2]$^t$). The control signal for the DDR robot (figure-13) may be constructed as (22)

$$\mathbf{U} = \begin{bmatrix} T_R \\ T_L \end{bmatrix} = \mathbf{Q_D^{-1}}(\mathbf{S} - \mathbf{S_D}). \tag{22}$$

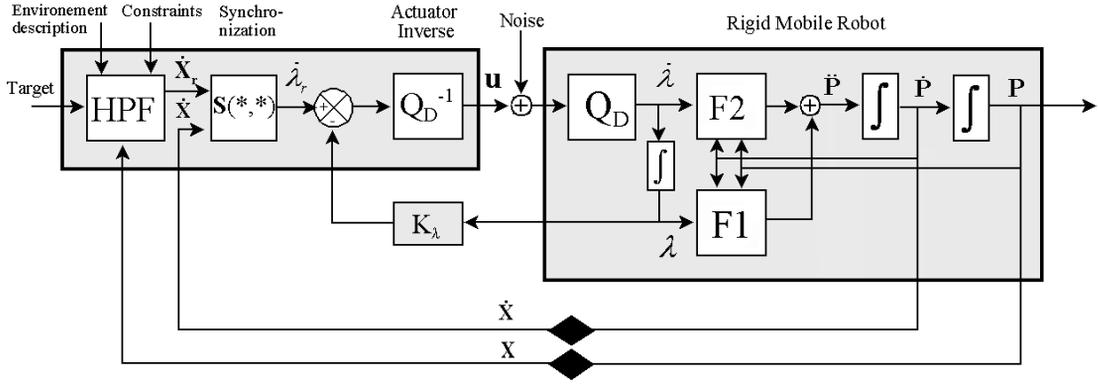

Figure-13: suggested planner for dynamic, separable mobile robots

## 6. Performance Analysis

In this section it is shown that the control approach suggested in section-5 can guarantee global asymptotic stability of the robots for the kinematic and dynamic cases. It is also shown for the kinematic case that the trajectory laid by the HPF planner and that of the robot can be made identical. In other words, the robot's trajectory inherits all the properties of the HPF trajectory.

It was proven in [61] that the gradient dynamical system in (2), which is constructed from an underlying harmonic potential, guarantees convergence from any point in $\Omega$ to a specified target point. The proof makes use of the fact that a harmonic potential is also a Liapunov function candidate. The LaSalle invariance principle [72], is used in the proof.

Proposition-1: The closed loop system constructed by using the control law in (16) with the system in (7) is stable and converges to the target position and orientation encoded in the harmonic potential.

Proof: Consider the Liapunov function candidate (23)

$$\Xi = V(x,y) + \frac{1}{2}(\Delta\theta)^2 \tag{23}$$

were $\Delta\theta = \arg(-\nabla V) - \theta$. Note that the HPF (V) is a Liapunov function that is positive everywhere in $\Omega$ except at $x=x_T$ and $y=y_T$ where it is equal to zero [61]. The derivative of $\Xi$ with respect to time is

$$\dot{\Xi} = \nabla V \begin{bmatrix} \dot{x} \\ \dot{y} \end{bmatrix} - \Delta\theta \cdot \dot{\theta}. \tag{24}$$

Combining (9), (15) and (16) we obtain

$$\begin{bmatrix} \dot{x} \\ \dot{y} \\ \dot{\theta} \end{bmatrix} = \begin{bmatrix} \cos(\theta) & 0 \\ \sin(\theta) & 0 \\ 0 & 1 \end{bmatrix} \mathbf{Q}(\mathbf{Q}^{-1}(\begin{bmatrix} K1 \cdot \cos(\Delta\theta) \cdot |-\nabla V| \\ K2 \cdot \Delta\theta \end{bmatrix})) \tag{25}$$

Substituting (25) in (24) we get

$$\dot{\Xi} = \nabla V \begin{bmatrix} \cos(\theta) & 0 \\ \sin(\theta) & 0 \end{bmatrix} \begin{bmatrix} K1 \cdot \cos(\Delta\theta) |-\nabla V| \\ -K2 \cdot \Delta\theta \end{bmatrix} + \begin{bmatrix} 0 & \Delta\theta \end{bmatrix} \begin{bmatrix} K1 \cdot \cos(\Delta\theta) |-\nabla V| \\ -K2 \cdot \Delta\theta \end{bmatrix}. \tag{26}$$

The gradient of V may be expressed as

$$\nabla V = |\nabla V| \cdot \begin{bmatrix} \cos(\arg(-\nabla V) + \pi) \\ \sin(\arg(-\nabla V) + \pi) \end{bmatrix} \tag{27}$$

substituting (27) into (26) we have

$$\dot{\Xi} = \begin{bmatrix} -|\nabla V| \cdot \cos(\Delta\theta) & 0 \end{bmatrix} \begin{bmatrix} K1 \cdot \cos(\Delta\theta) \cdot |-\nabla V| \\ -K2 \cdot \Delta\theta \end{bmatrix} + \begin{bmatrix} 0 & \Delta\theta \end{bmatrix} \begin{bmatrix} K1 \cdot \cos(\Delta\theta) \cdot |-\nabla V| \\ -K2 \cdot \Delta\theta \end{bmatrix} \tag{28}$$

or

$$\dot{\Xi} = -\begin{bmatrix} |\nabla V| \cdot \cos(\Delta\theta) & \Delta\theta \end{bmatrix} \begin{bmatrix} K1 & 0 \\ 0 & K2 \end{bmatrix} \begin{bmatrix} |-\nabla V| \cdot \cos(\Delta\theta) \\ \Delta\theta \end{bmatrix}. \tag{29}$$

Equation (15) may be used to compute the minimum invariant set of the system ($|\nabla V|=0$, $\Delta\theta=0$) to which the robot will converge.

Since it is proven that an HPF is Morse [63] convergence of $|\nabla V|$ to zero implies convergence of x and y to $x_T$ and $y_T$ respectively. Also convergence of $\Delta\theta$ to zero implies that the robot will converge to the orientation encoded by the HPF at $x_T$ and $y_T$. Since $\Delta\theta \to 0$

$$\lim_{t\to\infty}\theta(t) \to \arg(-\nabla V). \tag{30}$$

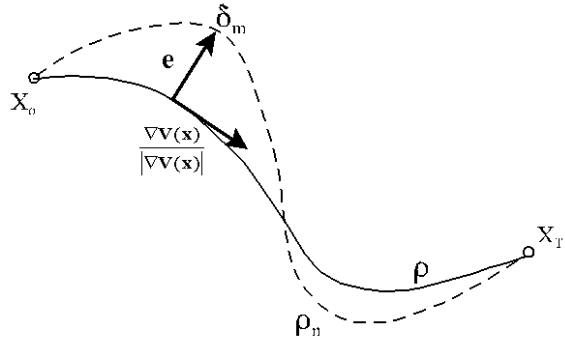

Figure-14: deviation between the HPF trajectory and the nonholonomic trajectory.

Proposition-2: Let $\rho$ be the spatial projection of the trajectory $\mathbf{X}(t)$ laid by the gradient dynamical system in (2). Also let $\rho_n$ be the spatial projection of the trajectory laid by the suggested nonholonomic planner (figure-14). Let $\delta(t)$ be the deviation between $\rho$ and $\rho_n$ at time t. Let $\delta_m$ be the maximum deviation. If the motion actuation stage of the robot is fully invertible, then $\delta_m$ may be made arbitrarily small,

Proof: Let $\mathbf{e}$ be a unit vector normal to $-\nabla V$ (31),
$$\mathbf{e} = \begin{bmatrix} \sin(\arg(-\nabla V)) \\ -\cos(\arg(-\nabla V)) \end{bmatrix}. \tag{31}$$

Also let $\Delta\theta = \arg(-\nabla V) - \theta$. The rate of change of $\delta(t)$ may be computed as (32)

$$\dot{\delta} = \mathbf{e}^{\mathbf{T}}\begin{bmatrix}\dot{x}\\\dot{y}\end{bmatrix} = \mathbf{e}^{\mathbf{T}}\begin{bmatrix}\cos(\theta) & 0\\ \sin(\theta) & 0\end{bmatrix}\begin{bmatrix} K1\cdot|-\nabla V|\cos(\Delta\theta)\\ K2\cdot\Delta\theta\end{bmatrix} \tag{32}$$

$$= \frac{K1\cdot|-\nabla V|}{2}\cdot\sin(2\cdot\Delta\theta)$$

We also have
$$\dot{\theta} = K2\cdot(\arg(-\nabla V)-\theta) = K2\cdot\Delta\theta \tag{33}$$

therefore
$$\frac{d\Delta\theta}{dt} + K2\cdot\Delta\theta = 0 \tag{34}$$

Solving the above equation (34) we have

$$\Delta\theta = \Delta\theta(0)\cdot exp(-K2\cdot t) \tag{35}$$

Equation-35 clearly shows that the velocity of the robot will converge in a controllable exponential manner to the reference velocity

$$\lim_{t\to\infty}\dot{\mathbf{X}}(\mathbf{X}(t)) \to \dot{\mathbf{X}}_r(\mathbf{X}(t)) \tag{36}$$

Since V is harmonic,
$$|\nabla V| \le C_m \qquad x,y \in \Omega \tag{37}$$

where $C_m$ is a finite positive constant [73]. Substituting (35) in (32), we have

$$\dot{\delta} = \frac{K2\cdot|-\nabla V|}{2}\sin(2\cdot\Delta\theta(0)\cdot\exp(-K2\cdot t)) \tag{38}$$
$$\le K2\cdot C_m\cdot\Delta\theta(0)\cdot\exp(-K2\cdot t)$$

Since the system is stable, we have
$$\delta(0) = 0, \quad \lim_{t\to\infty}\delta(t) \to 0, \tag{39}$$

$$\lim_{t\to\infty}\dot{\delta}(\infty) \to 0, \text{ and } \int_0^\infty \dot{\delta}(t)dt = 0$$

Since (38) satisfies the Lipschitz condition and depends mainly on (15), an upper bound on $\delta$ exists. From the above we conclude the following:

1 - If $\Delta\theta(0) = 0$ at t=0, then $\delta(t)=0 \ \forall t$.
In other words, the trajectory of the robot and the reference trajectory from the harmonic potential are identical. This means that all the provably-correct properties of a harmonic potenial field-generated trajectory can be migrated to the trajectory of the nonholonomic robot.

2- An upper bound on δ(t) (40) may be constructed as

$$\delta(t) \leq \frac{K1}{K2} C_m \cdot \Delta\theta(0) \cdot \exp(-K2 \cdot t) \leq \frac{K1}{K2} C_m \cdot \Delta\theta(0). \quad (40)$$

In other words, in the worst case, the deviation of the robot's trajectory from the kinematic trajectory can be made arbitrarily small by controlling the value of K2. In the above cases, it is possible to perfectly invert the actuation stage. If this is not possible, the pseudo inverse approach may be used to construct the inversion operator $\mathbf{Q}^{-1}$. With no loss of generality, if $\mathbf{Q}$ is differentiable the actuation stage maybe expressed in a linear form around an operating control point $\mathbf{Uo}$ (41)

$$\lambda = \mathbf{A}(\mathbf{Uo})\mathbf{U} \quad (41)$$

where $\mathbf{A}$ is a matrix that need not necessarily be full rank. In this case the inverse operator may be constructed as:

$$\mathbf{U} = \mathbf{A}^+(\mathbf{Uo})\lambda \quad (42)$$

$\mathbf{A}^+$ is the pseudo inverse of $\mathbf{A}$ [42].

Proposition-3: A matrix $\mathbf{B}$ constructed as the product of a matrix $\mathbf{A}$ by its pseudo inverse ($\mathbf{B}=\mathbf{A}^+\mathbf{A}$) is positive semi-definite ($\mathbf{A}$ is not full-rank).

Proof: by definition, the pseudo inverse of $\mathbf{A}$ is

$$\mathbf{A}^+ = \lim_{\delta \to 0}(\mathbf{A}^T\mathbf{A} + \delta \cdot \mathbf{I})^{-1}\mathbf{A}^T = \lim_{\delta \to 0}\mathbf{A}^T(\mathbf{A}^T\mathbf{A} + \delta \cdot \mathbf{I})^{-1}. \quad (43)$$

Let $\mathbf{M}=\mathbf{A}^T\mathbf{A}$, and $\mathbf{Z}=\delta \cdot \mathbf{I}$. Since $\mathbf{M}$ is symmetric and $\mathbf{Z}$ is positive definite, they may be jointly diagonalizable [24] into

$$\mathbf{M}=\mathbf{H}^T\mathbf{\Lambda}\mathbf{H} \text{ and } \mathbf{Z}=\mathbf{H}^T\mathbf{H}, \quad (44)$$

where $\mathbf{H}$ is an orthonormal matrix and $\mathbf{\Lambda}$ is a diagonal matrix containing the eigenvalues ($\beta_i$'s) of $\mathbf{M}$. Substituting (44) into (43) we have

$$\mathbf{B} = \lim_{\delta \to 0}(\mathbf{H}^T\mathbf{H} + \mathbf{H}^T\mathbf{\Lambda}\mathbf{H})^{-1}\mathbf{H}^T\mathbf{\Lambda}\mathbf{H} =$$
$$\lim_{\delta \to 0}[\mathbf{H}^T(\mathbf{I}+\mathbf{\Lambda})^{-1}\mathbf{\Lambda}\mathbf{H}] \quad (45)$$

$$= \lim_{\delta \to 0}[\mathbf{U}^T \begin{bmatrix} \frac{\beta_1}{1+\beta_1} & 0 & \cdot & 0 \\ 0 & \frac{\beta_2}{1+\beta_2} & \cdot & 0 \\ \cdot & \cdot & \cdot & \cdot \\ 0 & \cdot & 0 & \frac{\beta_N}{1+\beta_N} \end{bmatrix} \mathbf{U}]. \quad (46)$$

Since $\mathbf{M}$ is constructed as the product of a matrix by its transpose, its eigenvalues are non-negative. Therefore, the eigenvalues of $\mathbf{B}$ lie in the interval [0,1), i.e. they are non-negative. Therefore $\mathbf{B}$ is positive semi-definite. Similar treatment applies to $\mathbf{B}=\mathbf{A}\mathbf{A}^+$.

Proposition-4: The inverse-based control scheme suggested in (42) can guarantee a stable response for over and under-actuated robots when exact inversion is not possible.

Proof: Consider the Liapunov function in (23). If the pseudo inverse $\mathbf{A}^+$ of the actuation stage is used to generate the control signal, we have

$$\begin{bmatrix} \dot{x} \\ \dot{y} \\ \dot{\theta} \end{bmatrix} = \begin{bmatrix} \cos(\theta) & 0 \\ \sin(\theta) & 0 \\ 0 & 1 \end{bmatrix} \mathbf{B} \begin{bmatrix} K1 \cdot \cos(\Delta\theta) \cdot |-\nabla V| \\ K2 \cdot \Delta\theta \end{bmatrix} \quad (47)$$

where $\mathbf{B}=\mathbf{A}\mathbf{A}^+$. Substituting (47) in (24) we get (48)

$$\dot{\Xi} = \nabla V \begin{bmatrix} \cos(\theta) & 0 \\ \sin(\theta) & 0 \end{bmatrix} \mathbf{B} \begin{bmatrix} K1 \cdot \cos(\Delta\theta) |-\nabla V| \\ -K2 \cdot \Delta\theta \end{bmatrix} +$$
$$\begin{bmatrix} 0 & \Delta\theta \end{bmatrix} \mathbf{B} \begin{bmatrix} K1 \cdot \cos(\Delta\theta) |-\nabla V| \\ -K2 \cdot \Delta\theta \end{bmatrix}. \quad (48)$$

The gradient of V may be expressed as

$$\nabla V = |\nabla V| \cdot \begin{bmatrix} \cos(\arg(-\nabla V) + \pi) \\ \sin(\arg(-\nabla V) + \pi) \end{bmatrix} \quad (49)$$

substituting (49) into (48) we have

$$\dot{\Xi} = \begin{bmatrix} -|\nabla V| \cdot \cos(\Delta\theta) & 0 \end{bmatrix} \mathbf{B} \begin{bmatrix} k1 \cdot \cos(\Delta\theta) |-\nabla V| \\ -K2 \cdot \Delta\theta \end{bmatrix} + $$
$$\begin{bmatrix} 0 & \Delta\theta \end{bmatrix} \mathbf{B} \begin{bmatrix} K1 \cdot \cos(\Delta\theta) |-\nabla V| \\ -K2 \cdot \Delta\theta \end{bmatrix} \quad (50)$$

or
$$\dot{\Xi} = -\begin{bmatrix} |\nabla V| \cdot \cos(\Delta\theta) & \Delta\theta \end{bmatrix} \mathbf{B} \begin{bmatrix} K1 & 0 \\ 0 & K2 \end{bmatrix} \begin{bmatrix} \cos(\Delta\theta) |-\nabla V| \\ \Delta\theta \end{bmatrix},$$

there are three cases:

1- Redundant actuation (**A** has more columns than rows), for this case **B**=**I** (i.e. positive definite).
2- Full actuation (**A** is a square, full rank matrix), for this case **B**=**I** (i.e. positive definite).
3- Under actuation (**A** has more rows than column), for this case **B** is positive semi-definite.
The above means that, in the worst case, the suggested controller will be able to maintain stability. However, convergence to the target position and configuration cannot be guaranteed.

Proposition-5: The control law in (22) applied to a differential drive robot with second order dynamics described by the system equation in (11,12) guarantees global asymptotic convergence of the robot from any initial position and orientation in $\Omega$ to the target potion point $(x_T, y_T)$ and orientation $(\arg(-\nabla V(x_T, y_T)))$ encoded in the harmonic potential V for any positive K1, K2, KD1 and KD2 where KD1>K1.

Proof: consider the following Liapunov function candidate (51)
$$\Xi = K1 \cdot M \cdot V(x,y) + \frac{1}{2} K2 \cdot I \cdot (\Delta\theta)^2 + \frac{1}{2} I \cdot \dot{\theta}^2 + \frac{1}{2} M \cdot v^2 \quad (51)$$

where M is the mass of the robot and I is its inertia. Notice that V(x,y) is a valid liapunov function [61]. It is always positive except at the target configuration $(x_T, y_T, \theta_T)$ where it is equal to zero.

Let us use
$$X = \begin{bmatrix} x \\ y \end{bmatrix}, \qquad \dot{X} = \begin{bmatrix} \dot{x} \\ \dot{y} \end{bmatrix} = v \begin{bmatrix} \cos(\theta) \\ \sin(\theta) \end{bmatrix}. \quad (52)$$

The time derivative of $\Xi$ is:
$$\dot{\Xi} = K1 \cdot M \cdot \nabla V^t \dot{X} - K2 \cdot I \cdot \Delta\theta \cdot \dot{\theta} + I \cdot \dot{\theta} \cdot \ddot{\theta} + M \cdot v \cdot \dot{v} \quad (53)$$

since for a differential drive robot the inverse of the actuation stage always exist, the following substitutions may be used
$$\dot{v} = K1 \cdot (|-\nabla V| - v) \cdot \cos(\Delta\theta) - KD1 \cdot \eta_1(-\nabla V, \dot{X}) \cdot v \quad (54)$$
$$\ddot{\theta} = K2 \cdot \Delta\theta - KD2 \cdot \eta_2(-\nabla V, \dot{X}) \cdot \dot{\theta}$$

Substituting (52) and (54) in (53), the time derivative of $\Xi$ becomes (55)
$$\dot{\Xi} = -KD2 \cdot I \cdot \dot{\theta}^2 - M \cdot v^2 \cdot (KD1 \cdot \eta_1(-\nabla V, \dot{X}) + K1 \cdot \cos(\Delta\theta)). \quad (55)$$

Regardless of the choice of $\eta_1$, the time derivative of the Liapunov function is negative semi-definite if KD1 and KD2 are positive and KD1>K1. According to LaSalle principle, motion will converge to a subset of the set of points (E) for which the time derivative of $\Xi$ is zero
$$E = \{\dot{\rho} = 0, \dot{\theta} = 0, x, y, \theta\}. \quad (56)$$

The minimum invariant set ($\chi$) may be computed as the subset of E for which the gradient dynamical system in (2) vanishes. It was shown in [61] that motion for (2) is guaranteed to converge to the target point $x_T, y_T$, hence the orientation of the robot will converge to $\arg(-\nabla V(x_T, y_T))$. The dynamical differential drive robot will converge to the set
$$\chi = \{v = 0, \dot{\theta} = 0, x = x_T, y = y_T, \theta = \arg(-\nabla V(x_T, y_T))\} \quad (57)$$
provided that K1 and K2 are positive.

By placing $\eta_1$ in the selective tangential velocity damping mode (21) and selecting a high value of KD1, the maximum deviation between the robot's trajectory and the trajectory laid by the HPF planner ($\delta_m$) can be kept below a given arbitrary small value. This is due to the fact that motion will be severely impeded unless the velocity vector of the robot is aligned with the reference velocity vector from the gradient of the harmonic potential.

## 7. Simulation Results

This section presents several simulation examples designed to explore the properties of the suggested scheme and demonstrate its utility.

## 7.1 Intuitive parameter tuning (Kinematic case):

This example demonstrates two things. The first is the controller's ability to efficiently perform point-to-point movements. The second one is the ease in which the controller's parameters can be tuned. An FSR robot is required to move in free space from an initial position (0,-1) to a final position (1,0). The initial orientation of the robot is $\pi/2$. The gradient guidance field is shown in figure-15.

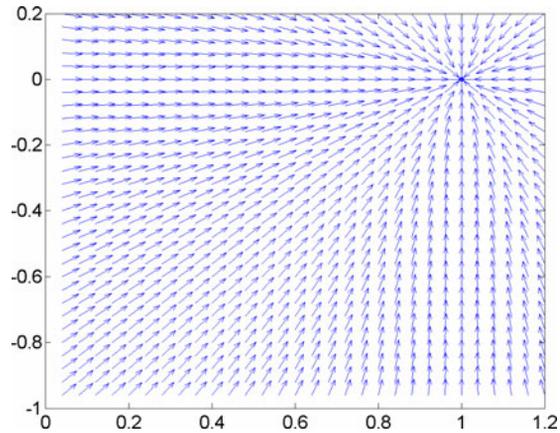

Figure-15: The guidance field in a free space environemnt with target point at x=0 and y=1.

The controller's parameters are initially selected as K1=K2=4. As can be seen from figure-16 the robot spirals towards the target in a manner that is akin to an under damped response. The x and y components of the trajectory are shown in figure 17. The corresponding, well-behaved control signal is shown in figure-18.

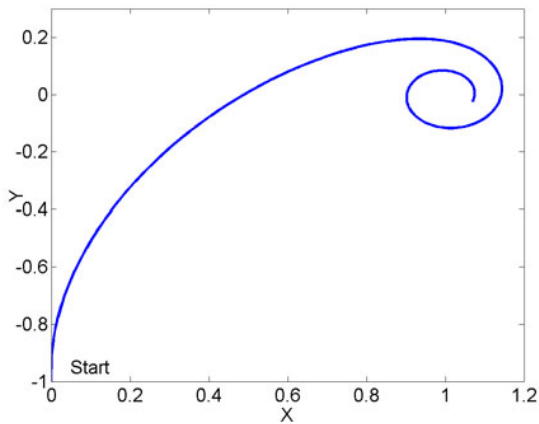
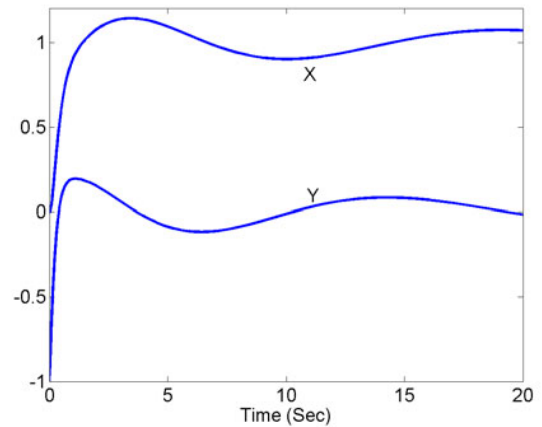

Figure-16: point-to-point spatial trajectory of a kinematic FSR robot.    Figure-17: The x & y components corresponding to the trajectory in figure-16

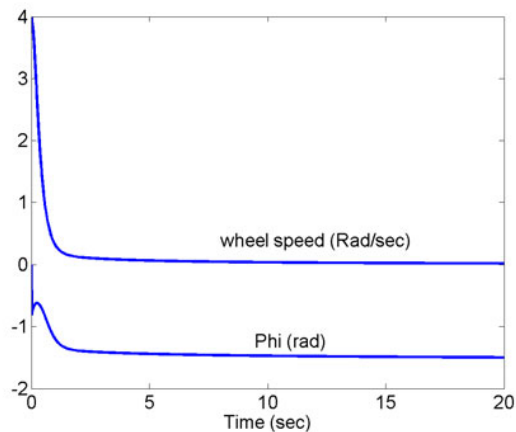

Figure-18: control signal of a kinematic FSR robot (K1=K2=4) corresponding to figure-16

Tuning a nonlinear controller is not an easy thing to do. However, the parameters of the suggested controller (K1 and K2) can be directly tied to the projected behavior. This greatly simplifies the tuning process. To understand such a relation one must notice that K1 controls the robot's tangential speed while K2 is responsible for aligning the robot's velocity with the desired reference velocity. A high K1 is likely to cause the overshoot behavior while a high value of K2 results in quick alignment of the dynamical behavior of the robot with that encoded in the reference guidance field. This is tested by reducing the value of K1 in the previous example to K1=1 while keeping K2=4. Figure-19 shows the robot's trajectory and figure-20 shows the corresponding x and y components as a function of time. As can be seen the oscillations disappeared and the robot converged to the target in a steady over-damped manner. The well-behaved control signals are shown in figure-21.

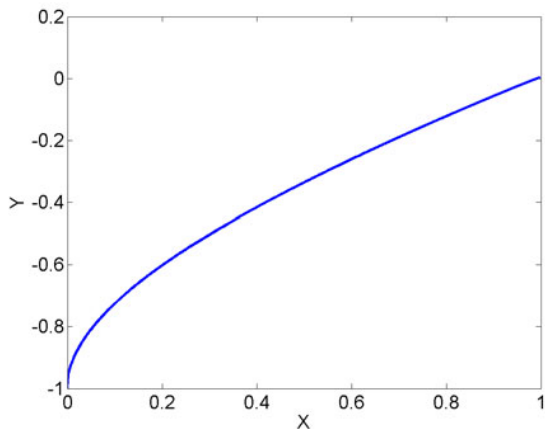
Figure-19: spatial trajectory of a kinematic FSR robot, K1=1, K2=4.

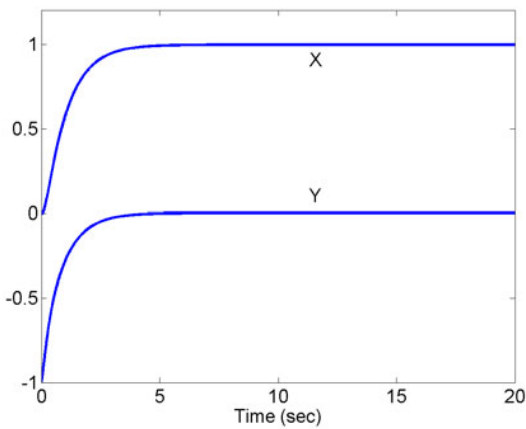
Figure-20: x & y components of the trajectory in figure-19.

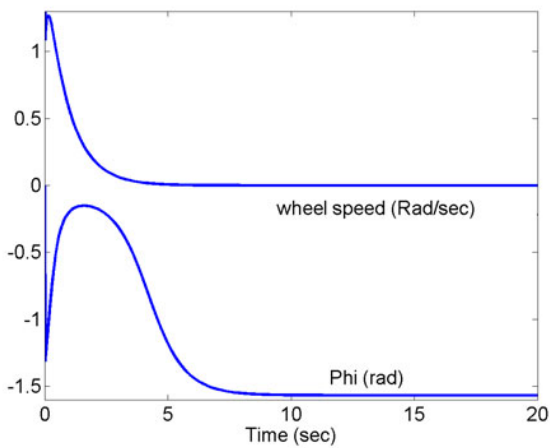
Figure-21: control signal corresponding to the trajectory in figure-19.

## 7.2 Guidance tracking (Kinematic case):
In this example the ability of the controller to make an FSR robot track a behavior that is encoded in a guidance field is demonstrated. A free space environment is used. The initial conditions are the same as in example 7.1. The controller parameters are kept the same as in the previous example (K1=1, K2=4). The gradient guidance field is shown in figure-22. It encodes the behavior: move along the x-axis at a constant speed and maintain the centerline.

The generated trajectory is shown in figure-23 and its x and y components are shown in figure-24. As can be seen the trajectory converges in a well-behaved manner to the desired trajectory with constant x-velocity and zero y-velocity. The corresponding control signals (figure-25) are also well-behaved.

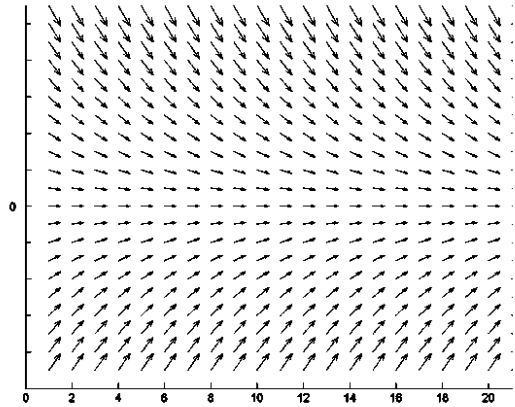

Figure-22: Move along the x-axis and stay at center gradient guidance field.

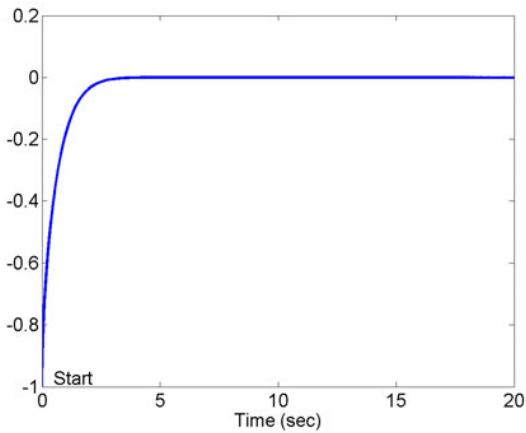

Figure-23: spatial trajectory of a FSR robot, centerline motion along x.

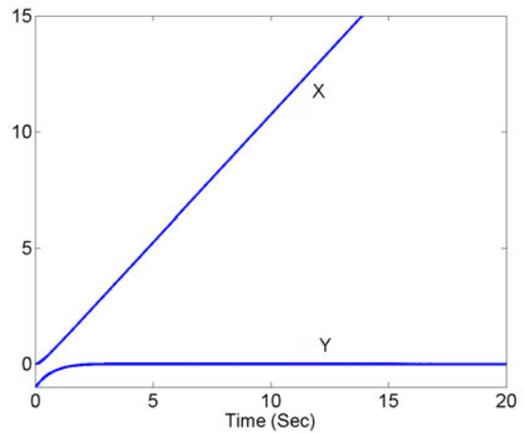

Figure-24: x & y components of the trajectory in figure-23.

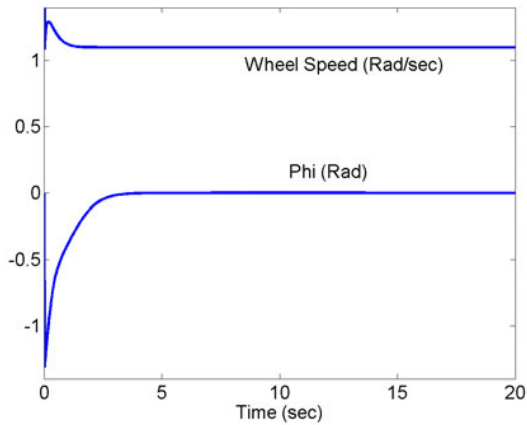

Figure-25: control signal of a the trajectory in figure-23

## 7.3 Robustness assessment (Kinematic case):

The robustness of the controller is tested for both actuator noise and saturation. The environment, navigation field and controller used in this example are the same as the ones used in example 7.2. A uniformly-distributed noise (-.2, .2) is added to the control signal corresponding to the trajectory shown in figure-23. The control signal is then subjected to a saturation nonlinearity that limits the value of the control signal to the period (-.5, .5) before applying it to the robot. The resulting trajectory is shown in figure-26 and the corresponding x and y components are shown in figure-27. The control signals are shown in figure-28. Despite the severity of the disturbance, the robot is still able to execute the behavior in the guidance field and move at a constant velocity along the centerline. The transient response was marginally affected and the path remains smooth.

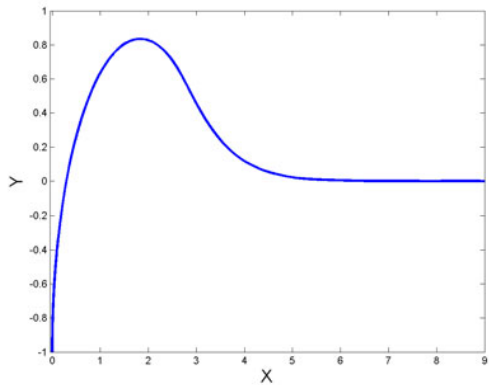
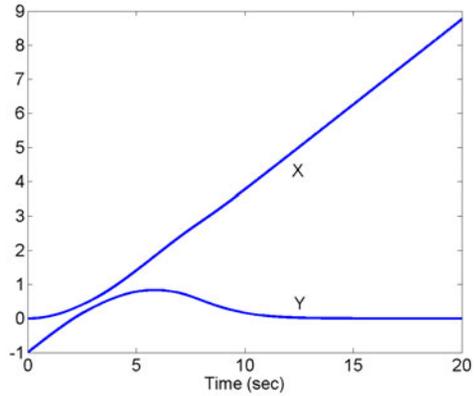

Figure-26: spatial trajectory an FSR robot with actuator saturation and noise    Figure-27: x & y components corresponding to the trajectory in figure-26

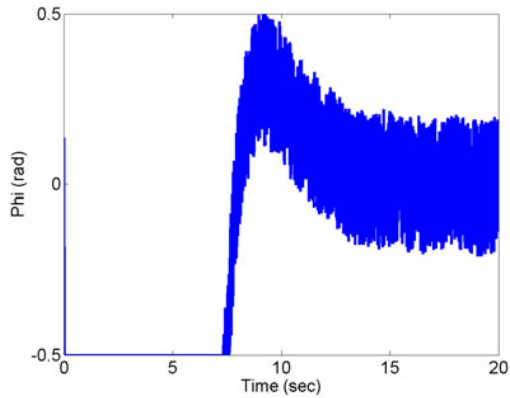
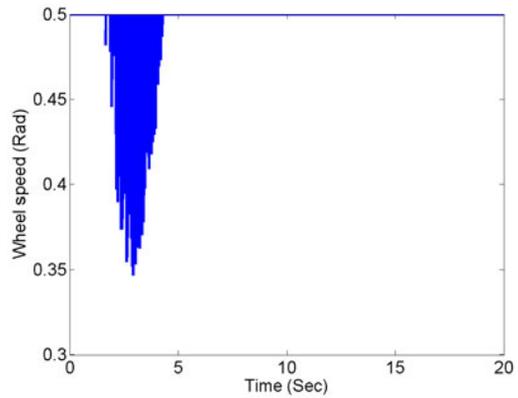

Figure-28: control signals of an FSR corresponding to the trajectory in figure-26

The example in 7.2 is used to test the robustness of the controller to uncertainty in the parameters. The actual FSR robot length is L=1 and the driving wheel radius is r=1. The parameters used by the controller are L=1/2 and r=1.5. The resulting trajectory is shown in figure-29 and the control signals are shown in figure-30. As can be seen, the robot is still able to actualize the guidance signal. The impact of the parameter error on the transients is quite marginal and the control signals are still well-behaved.

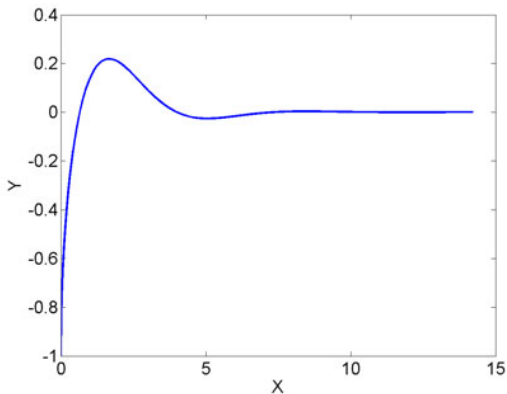
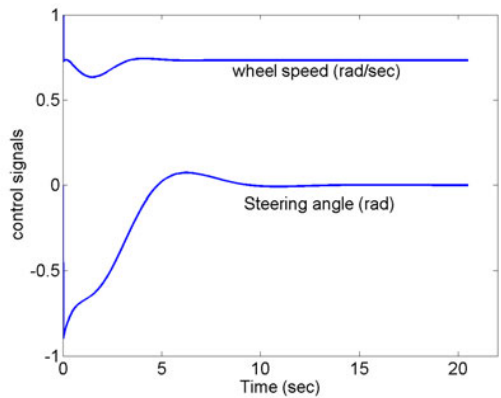

Figure-29: trajectory of the robot in figure-23 with parameter errors    Figure-30: control corresponding to the trajectory in figure-29.

### 7.4 Navigating a complex environment (Kinematic case):
In this example, the ability of the controller to make an FSR robot navigate a complex environment is demonstrated. The environment is described by a scalar field (uncertainty map) representing the fitness of each point of space to support motion (figure-31). The same controller used in example 7.2 is used with this example. The guidance field (figure-32) is generated using the gamma-harmonic approach (GHPF) [65,66] described earlier in the paper (equation-6). The resulting trajectories are shown in figure-33. Both trajectories are superimposed on an intensity map that represents the fitness field describing the environment. The higher the brigness of the intesity map the better support the point in space have for motion. The dotted blue trajectory is laid by the GHPF planner and the solid red trajectory is laid by the robot. By aligning the initial orientation of the robot's trajectory with that of the trajectory generated by the GFPF planner, both the robot's path and the GHPF planner path can be made almost identical. As can be seen, instead of moving along a straight line to the target point, the trajectory moves mainly in bright areas. This is an indicator that the trajectory maximize the overall mission success. It is also worth noting that despite the fact that the environment fitness descriptor is not smooth, both paths from the GHPF planner and the robot are smooth.

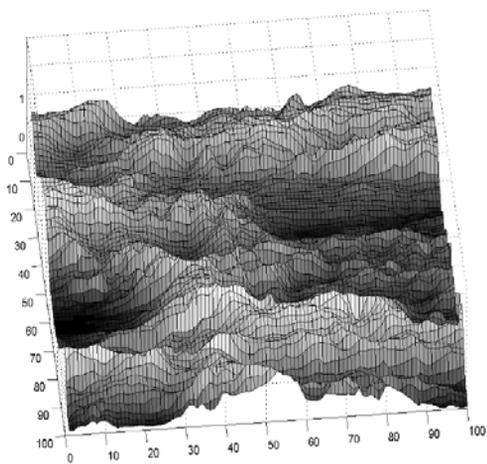
Figure-31: An environment described by an uncertainty map

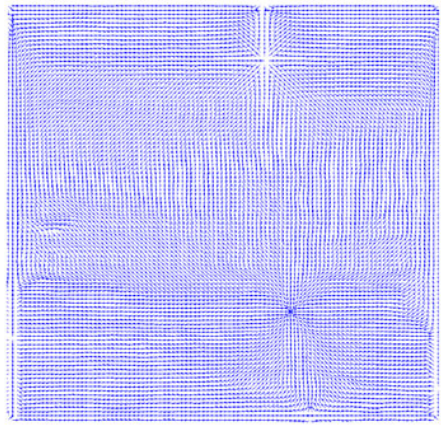
Figure-32 The guidance field corresponding to figure-31.

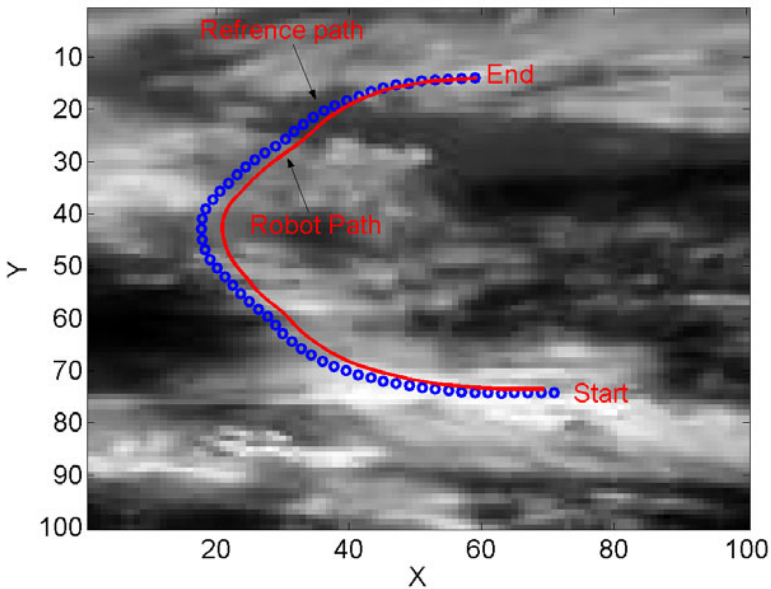
Figure-33: FSR robot navigating in an environment described by the uncertainty map in figure-31.

### 7.5 Kinematic controller failure with dynamic robots
This example demonstrates that the controller designed for the kinematic robot will not be able to stabilize a dynamic robot. A free space environment is used. The same guidance field in example 7.2 (figure-22) is used with a DDR robot. The controller's parameters are K1=1, K2=4. The robot starts from the initial position (0,-1) and the initial orientation $\theta=\pi/2$. Figure-34 shows the trajectory obtained by applying the controller to the kinematic robot. Figure-35 shows the x and y components corresponding to the trajectory. The well-behaved control signals are shown in figure-36. As can be seen, the controller efficiently managed to convert the guidance field into a control signal. Figure-37 shows the response of the kinematic controller when applied to the dynamic DDR robot. As can be seen, the response is unstable.

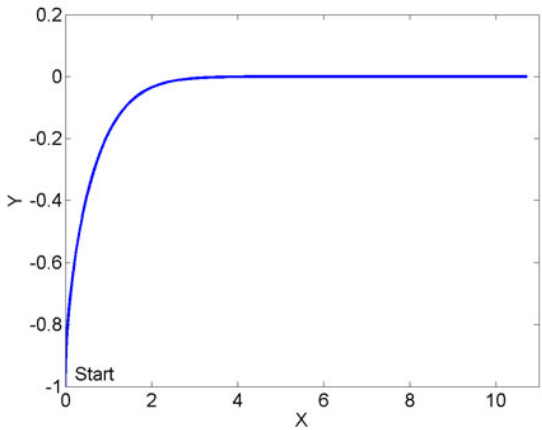
Figure-34: trajectory of a kinematic DDR robot corresponding to the guidance field in figure-22.

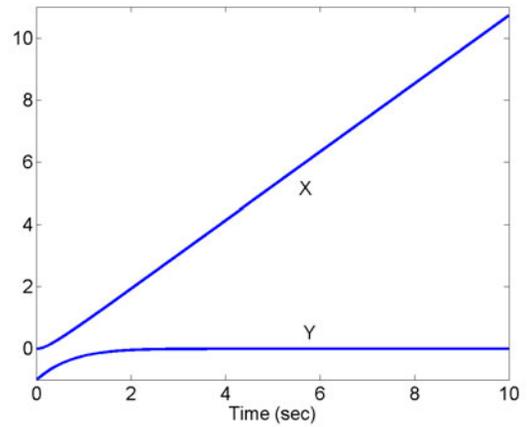
Figure-35: x & y components corresponding to figure-34.

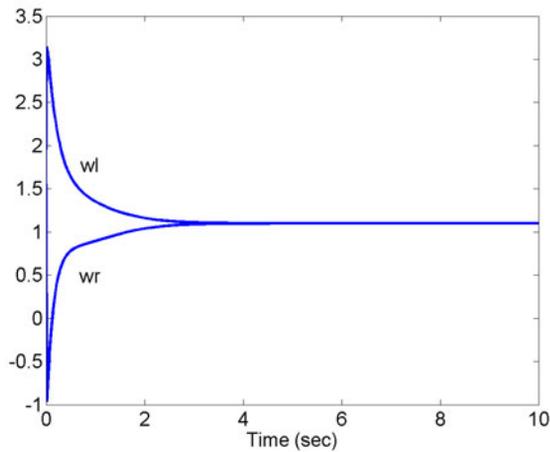
Figure-36: Control signal of a kinematic DDR robot corresponding to the trajectory in figure-34.

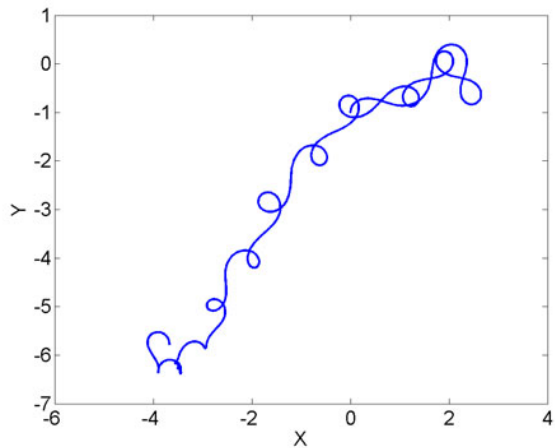
Figure-37: xy spatial trajectory, synchronizing field from figure-34 directly applied to the dynamical DDR robot, KD1=KD2=0

## 7.6 A basic example (Dynamic case):
The example in figure-7.5 is repeated using the dynamic controller suggested in equation-22. The parameters of the controller are: K1=1, K2=4, ($\eta_1=\eta_2=1$) with KD1=KD2=2. The DDR robot parameters are W=1 and r=1. The resulting trajectory is shown in figure-38 and corresponding x and y components are shown in figure-39. As can be seen, the addition of damping in the local coordinates of the robot stabilized the system and produced a well-behaved response.

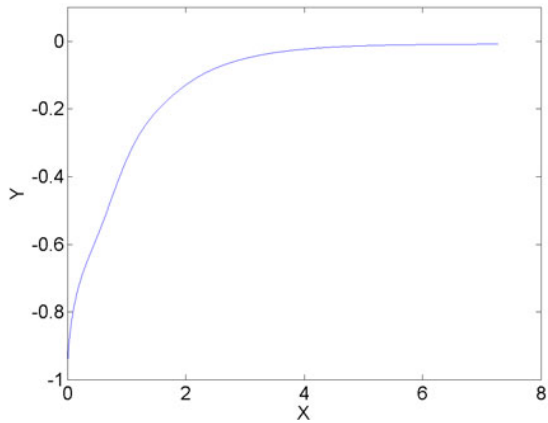

Figure-38: xy trajectory, damping increased to KD1= KD2=2 , $\eta_1=\eta_2=1$.

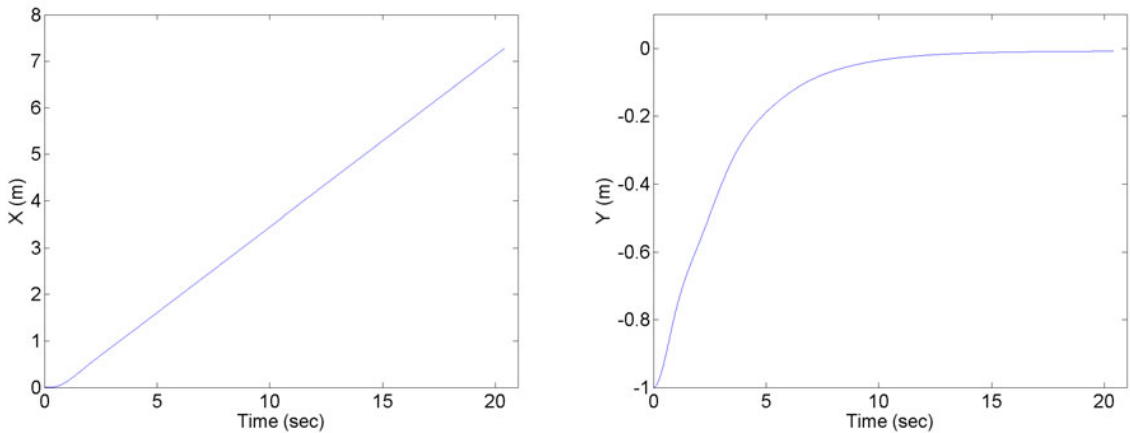

Figure-39: x vs time and y vs time corresponding to figure-38.

## 7.7 Directional damping added (Dynamic case):

This example demonstrate the ability of directionally-selective damping to yield a better response compared to the omni-directional damping. The same example in 7.6 is repeated; however, directional damping is used. All the parameters of the controller are kept the same as in example 7.6. The trajectory is shown in figure-40 and corresponding x and y components are in figure-41. The orientation and the tangential speed are in figure-42 and the well-behaved torques that are controlling the wheels of the robot are shown in figure-43. Directional damping significantly enhanced the robot's response increasing its rate of convergence by almost a factor of three with negligible oscillations in the trajectory. All attributes of the trajectory and the control signals are well-behaved.

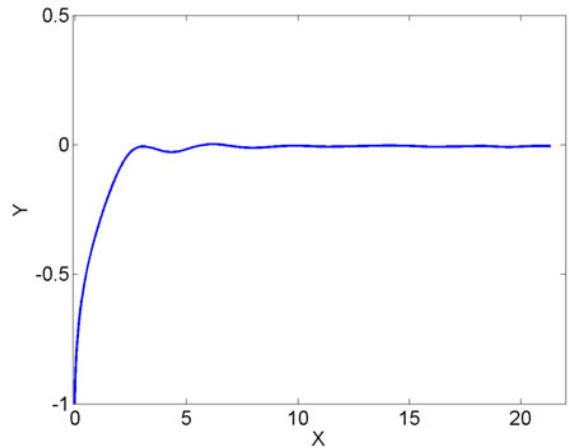

Figure-40: trajectory, KD1= KD2=2 , selective damping used.

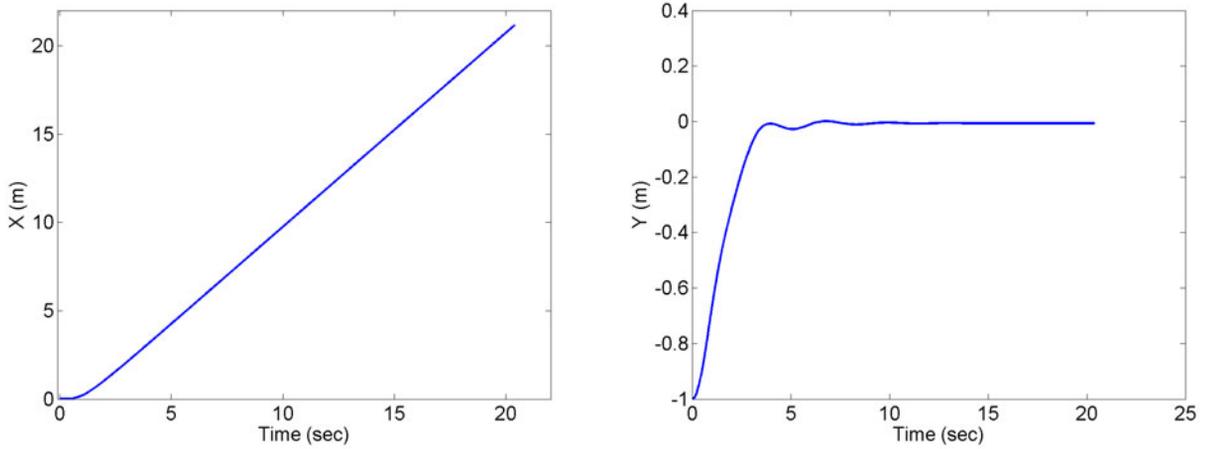
Figure-41: xy trajectory, x vs time and y vs time, corresponding to figure-40.

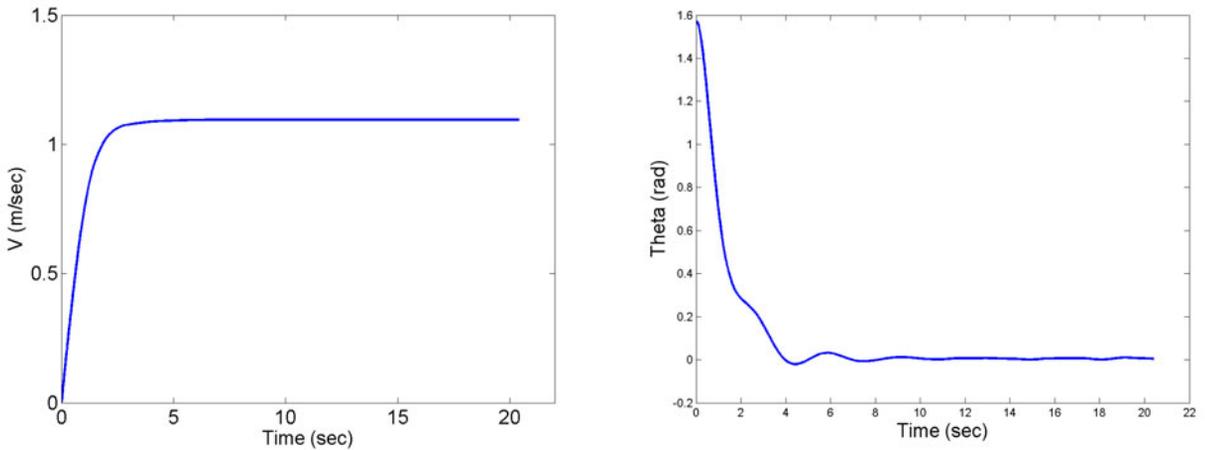
Figure-42: tangential and angular velocities associated with figure-40

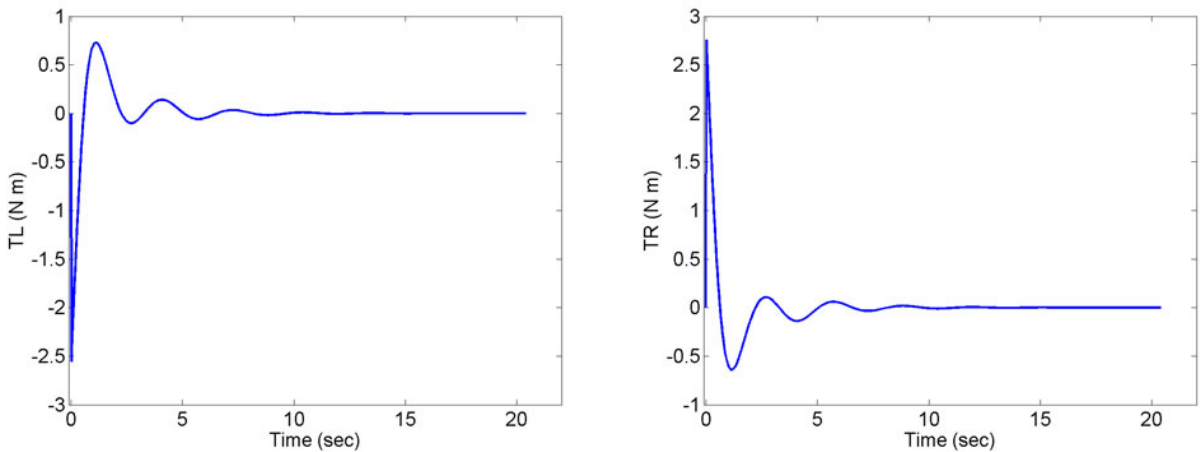
Figure-43: control torques associated with trajectory in figure-40

## 7.8 Motion with antipodal guidance (Dynamic case)

This example tests the behavior of the controller when guidance exhibit an antagonist action on the robot's motion. Example-7.7 is repeated with $\theta(0)=-\pi/2$. This places the robot in an antipodal heading that is diametrically opposite to direction of the guidance field. Figure-44 shows the trajectory of the robot and figure-45 shows the corresponding x and y components. The tangential velocity and the orientation of the robot are shown in figure-46. As can be seen, having components of the guidance field acting in a direction opposite to the robot's motion does not affect the robot's ability to generate a steady state motion that is in accordance with the guidance field. Moreover, the controller dealt with this situation while treating space as a scarce resource and performing the natural maneuver of backing (negative tangential velocity) while adjusting the orientation to become in-phase with the guidance. All attributes of the trajectory are well-behaved as well as the control signals that are actuating motion (figure-47).

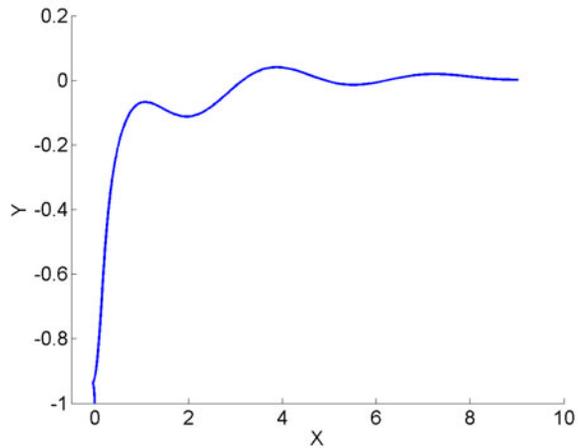

Figure-44: trajectory, x and y components, re-orientation maneuver.

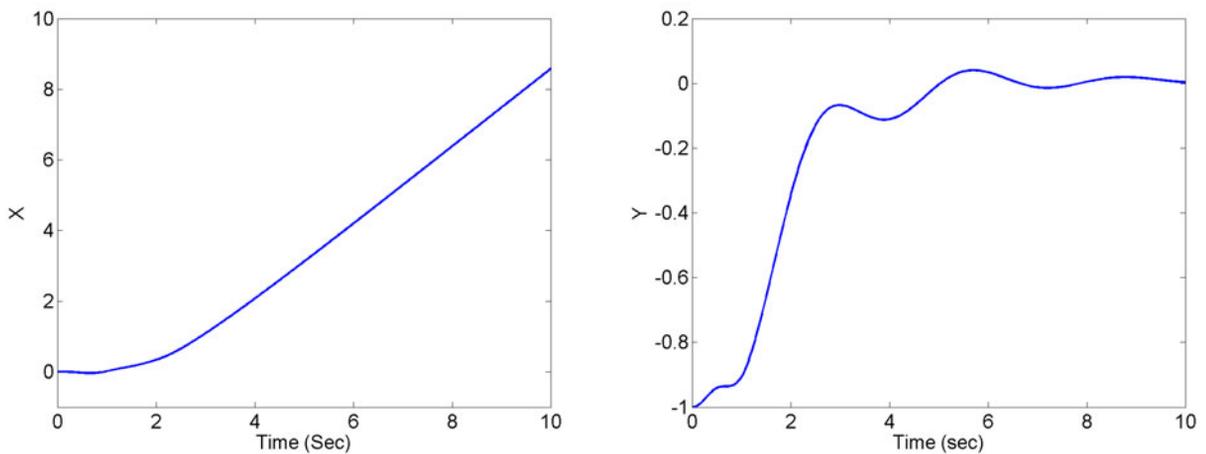

Figure-45: trajectory, x and y components, re-orientation maneuver.

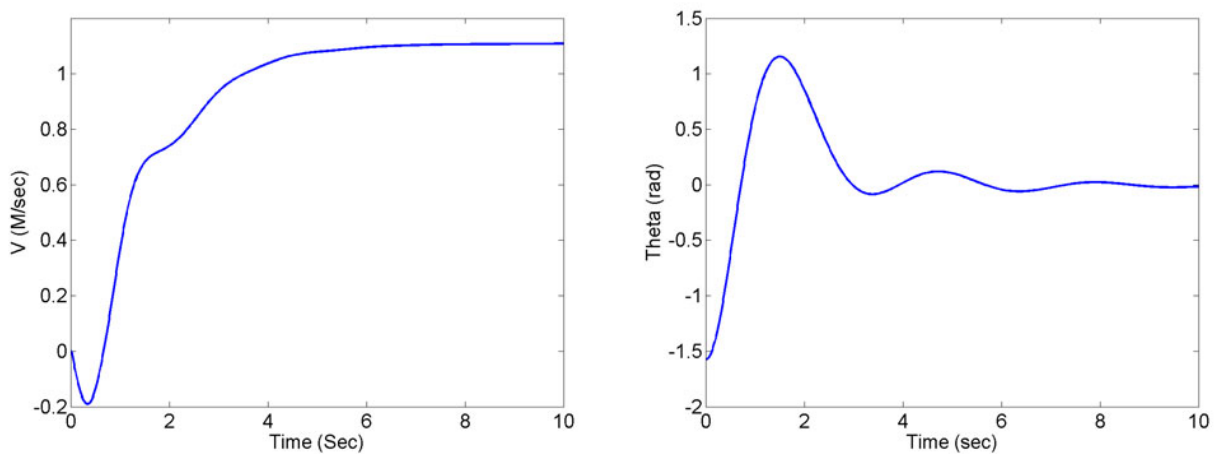

Figure-46: tangential velocity and orientation, re-orientation maneuver.

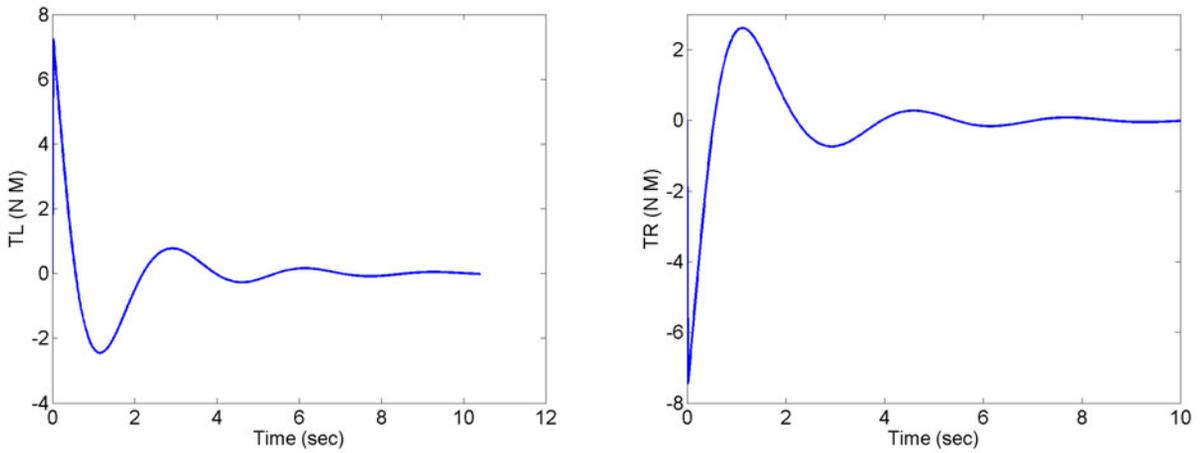
Figure-47: control signals, re-orientation maneuver.

## 7.9 Robustness assessment (Dynamic case):
This example tests the robustness of the dynamical controller. The same settings for the environment, robot and contorl as in example-7.7 are used. First, the robustness of the controller is tested in the presence of actuator noise by adding uniform noise in the period {-.5, .5} to the control signal. The resulting trajectory is shown in figure-48 and the control signals are in figure-49. As can be seen, the impact of noise on the response is minimal.

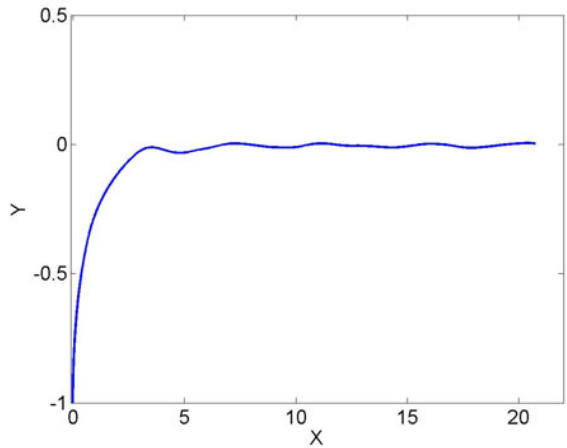
Figure-48: Trajectory. Same as figure-40 with actuator noise added

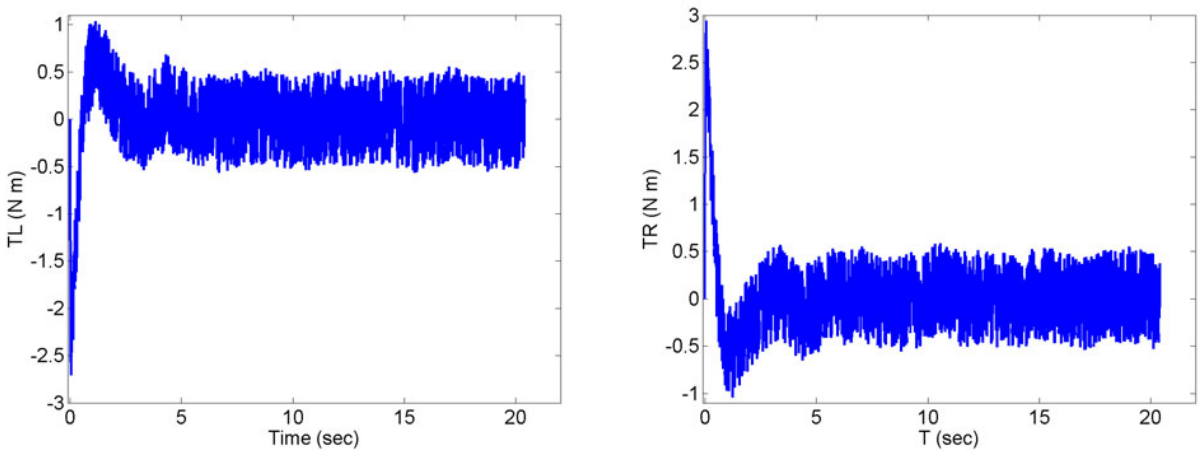
Figure-49: control signals corresponding to the trajctory in figure-48

The test is now repeated to assess the ability of the controller to tolerate actuator saturation. The control signal is restricted to the period {-.5, .5}. When compared to saturation-free maximum value of 2.8, this amounts to about 85% of the signal magnitude. The trajectory is shown in figure-50 and the control signals are in figure-51. As can be seen, other than slowing the response a little, the impact of this severe saturation nonlinearity is almost negligible. Comparable behavior was observed for high level of saturation. Abrupt change in the quality of the response is noticed at a 95% saturation {-.15, .15}. The trajectory for the 95% satuarion nonlinearity is in figure-52 and the corresponding control signal is in figure-53.

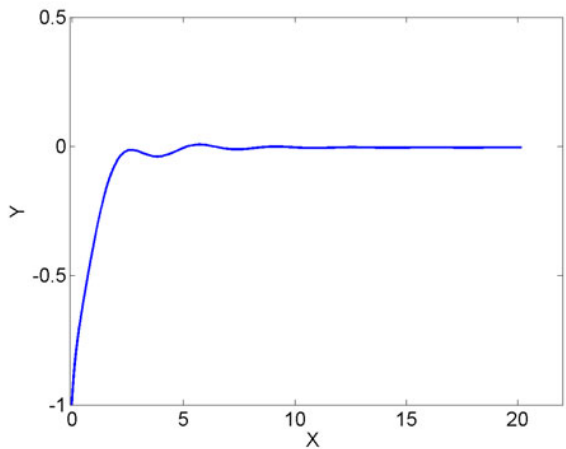

Figure-50: trajectory resulting from 85% actuator saturation

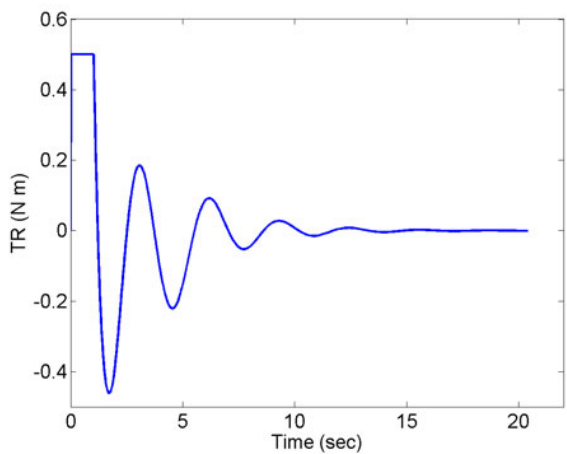
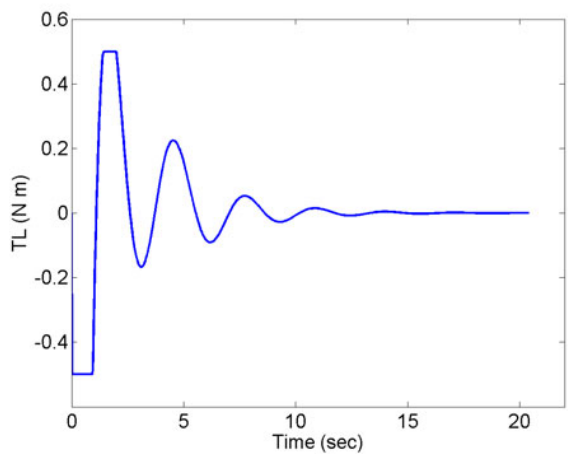

Figure-51: control signals, 85% actuator saturation

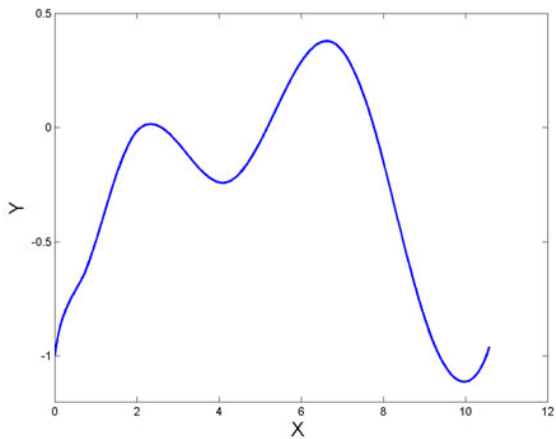

Figure-52: trajectory resulting from 95% actuator saturation

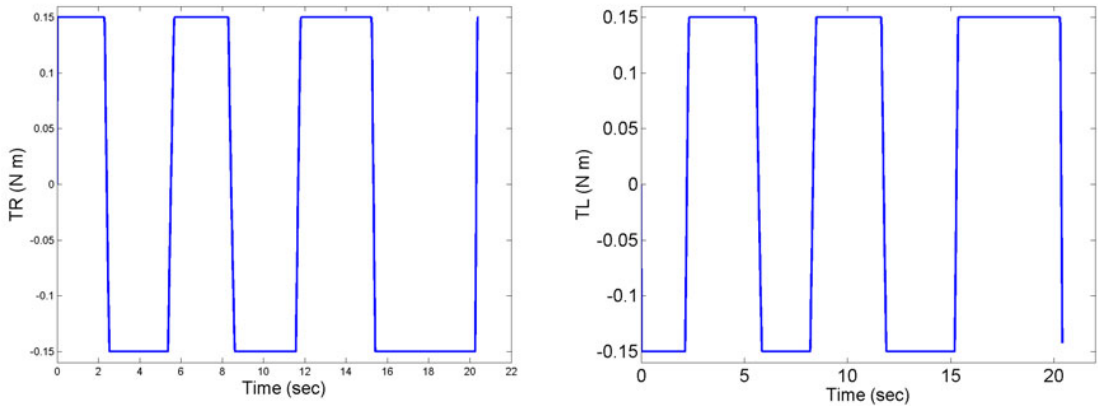

Figure-53: control signals., 95% actuator saturation

## 7.10 Compliance with the guidance and robustness assessment (Dynamic case):

This example tests the point-to-point motion capabilities of the kinodynamic controller for an involved harmonic guidance field. The HPF gradient guidance field superimposed on the environment is shown in figure-54. The kinematic trajectory is also shown in the figure. The initial conditions used are: $x(0)=y(0)=10$, $\theta(0)=0$. The final position is $x=y=40$ with no restrictions on orientation. The same controller used in example-7.7 is used in this case. Figure-55 shows the dynamic trajectory of the DDR robot. As can be seen, the HPF trajectory and the robot's trajectory are almost identical. The control signals are shown in figure-56. As can be seen the control signals are well-behaved.

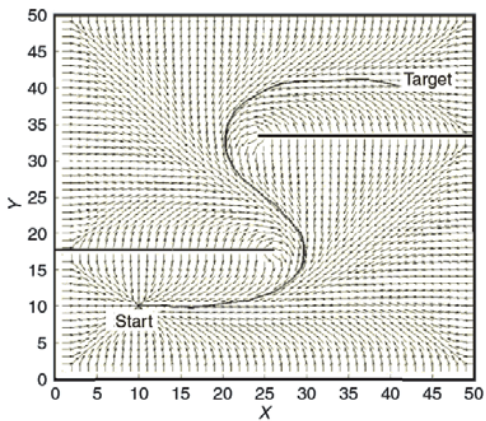
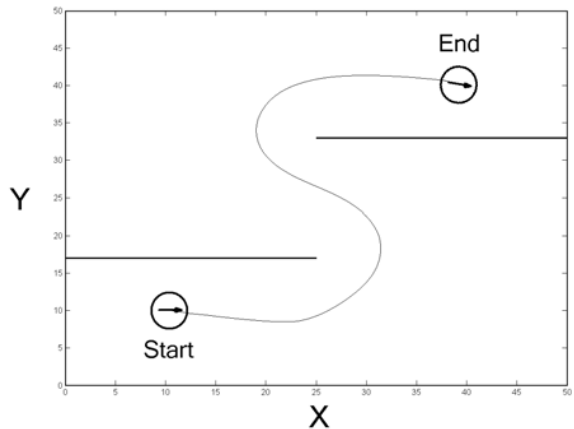

Figure-54: Guidance field and trajectory of an, HPF planner    Figure-55: Trajectory of the dynamical DDR robot corresponding to figure-54

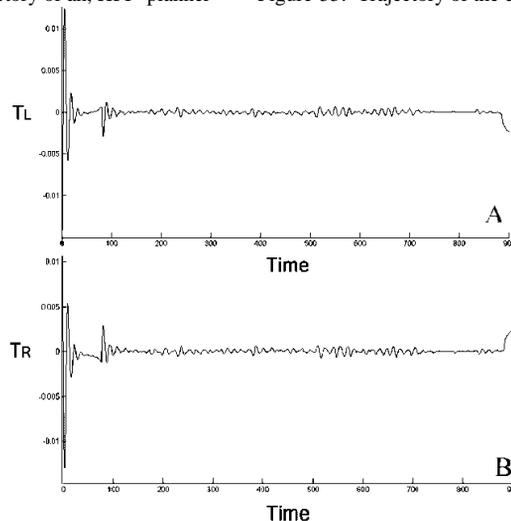

Figure-56: control torques corresponding to the trajectory in figure-55.

The robustness of the proposed controller in the presence of actuator saturation is tested. The magnitude of the torques ($T_R$ and $T_L$) is restricted not to exceed $T_m$ where

$$T_m = C \cdot \max(\max_t(|T_R(t)|), \max_t(|T_L(t)|)) \tag{58}$$

$T_R$ and $T_L$ are the torques for then non-saturated case, C is a constant representing the percentage saturation. The maximum torque for the non-saturated actuators is equal to .103 Nm. The controller showed remarkable robustness to saturation. The trajectory(figure-57) was virtually unaffected up to 99.8% saturation (i.e. C=.002). A sudden breakdown in performance is observed beyond this limit.

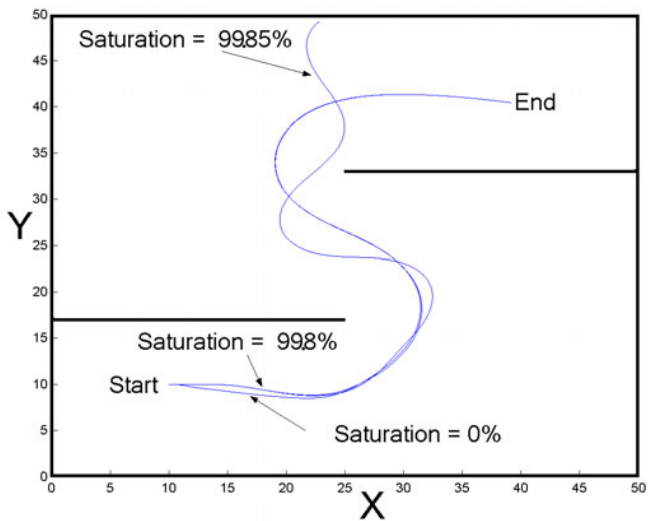

Figure-57: different xy trajectories in the presence of actuator saturation, same controller in figure-55

The same controller above is used with a more complex environment. The initial and final conditions are the same as above. The gradient guidance field is shown in figure-58. The trajectory is in figure-59. Same as in the previous case a well-behaved response is obtained.

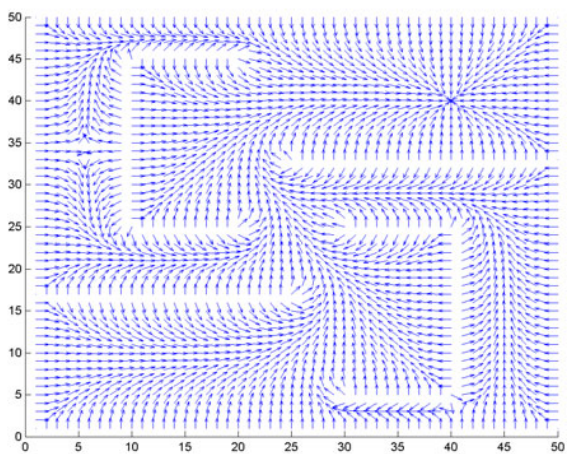
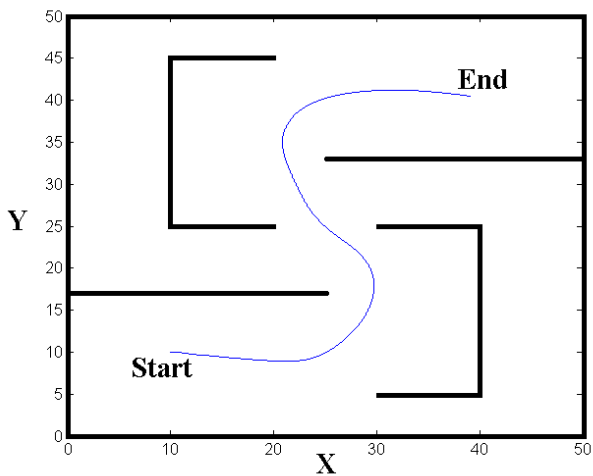

Figure-58: guidance HPF field and trajectory,    Figure-59: trajectory corresponding to the guidance field in figure-58.

## 7.11 Decentralized multi-agent planning (Kinematic case):

An important property of the HPF planning approach is the ability of the approach to act as a planning protocol. This means that two or more robots using an HPF planner can share the resource of space in a conflict-free manner. The ability of the suggested control to migrate such a property to a robot is tested using two FSR robots. The guidance field used for the first robot is similar to the one in figure-22. The second robot uses a similar guidance field; however, the field drives motion in the negative x direction. The initial conditions for the first robot are x(0)=-8, y(0)=-1 and θ(0)=π/2. The initial conditions for the second robot are x(0)=8, y(0)=1 and θ(0)=-π/2. Both controllers uses the same parameters as in example 7.2. The vector-harmonic potential field method suggested in [69] is used as the conflict resolving field. The trajectories of the robots are shown in figure-60. The distance between them as a function of time is shown in figure-61 and the control signals for each robots is shown in figure-62. Although the robots were faced with an immanent conflict that they were not aware of, they managed to collectively resolve it without exchanging intension and generate well-behaved, conflict-free trajectories that allows each robot to enforce the behavior encoded in the corresponding guidance field. The control signals that generated these trajectories are well-behaved.

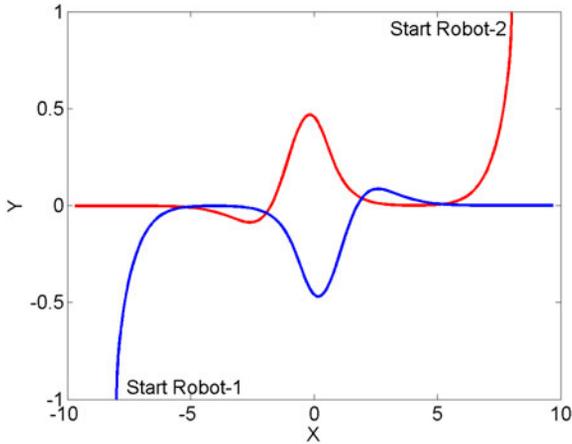
Figure-60: collision-free spatial trajectories for two FSR robots.

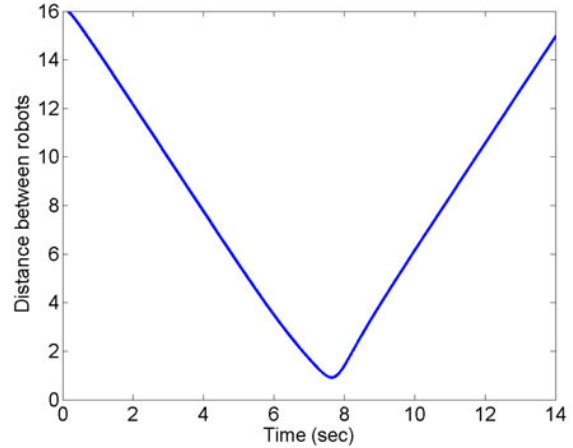
Figure-61: distance between the robots as a function of time

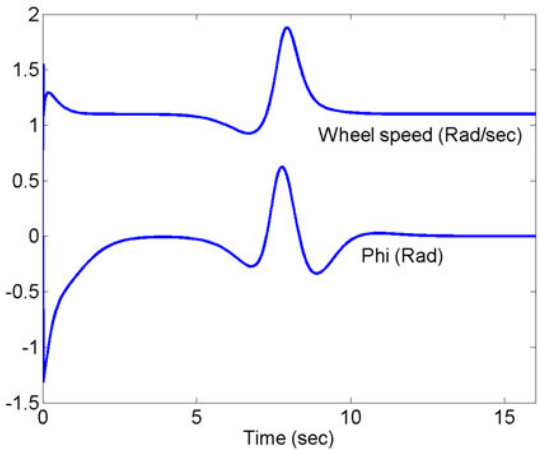

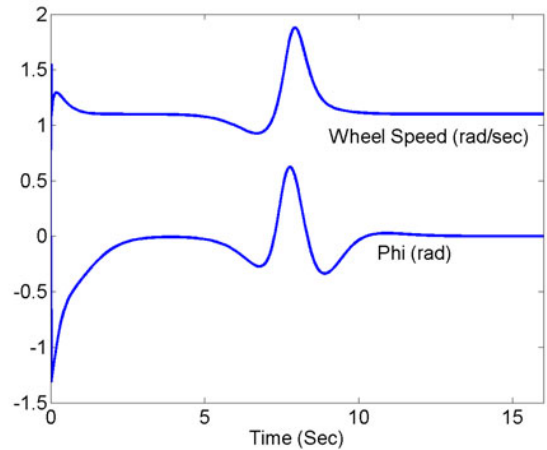

Figure-62: the control signals of the two FSR robots in figure-60.

## 8. Conclusions

This paper demonstrates the ability of the harmonic potential field motion planning approach to deal with realistic planning problems such as the kinodynamic motion planning of practical UGVs. Although the suggested solution is relatively simple (compared to the existing approaches) it encompasses several important features. The structure of the controller is simple, making it highly possible to implement using inexpensive hardware. Despite this simplicity, the controller can tackle the exact model of a separable UGV and provide an unconditionally stable, easy to tune response. The ability to effectively migrate, in a provably-correct manner, the kinematic path characteristics from an HPF-based planner to the dynamic trajectory of a UGV makes it possible to impose a wide class of constraints. It also allows different formats of data representing the environment to be used in generating the robot's motion. These two features along with the remarkable robustness the approach exhibits in the presence of actuator noise and saturation makes the suggested controller an effective candidate for addressing the challenges a realistic environment presents a robot with. The HPF approach is not only a planner, it is also a planning protocol. This enables a group of robots equipped with the same planner to share in a decentralized, conflict-free manner a common workspace. As demonstrated, the suggested approach is also able to preserve this valuable feature of the HPF approach.

**Acknowledgment:** the author acknowledges King Fahd University of Petroleum and Minerals for its support